\documentclass[preprint,12pt]{elsarticle}

\usepackage{amssymb}
\usepackage{hyperref}
\usepackage{algorithm}
\usepackage{algorithmic}
\usepackage{amsmath}
\usepackage{algorithm}
\usepackage{algorithmic}
\usepackage{graphicx}
\usepackage{subcaption}
\usepackage{mathtools}
\usepackage{amsthm}
\usepackage{bm}
\usepackage{booktabs} 
\usepackage{multirow} 
\journal{
Knowledge-Based Systems}

\begin{document}

\begin{frontmatter}



\title{Fully Bayesian Differential Gaussian Processes through Stochastic
Differential Equations}
\tnotetext[1]{The work is supported by the Fundamental Research Program of Guangdong, China, under Grant 2023A1515011281; and in part by the National Natural Science Foundation of China under Grant 61571005.}

\author[1]{Jian Xu}

\ead{2713091379@qq.com}



\affiliation[1]{organization={South China University of Technology},
    city={Guangzhou},
    country={China}}

\author[1]{Zhiqi Lin}
\ead{202311089192@mail.scut.edu.cn}

\author[1]{Min Chen}

\ead{minchen@ieee.org}


\affiliation[2]{organization={Columbia
University},
    city={New York},
    country={USA}}
\author[1]{Junmei Yang}
\ead{yjunmei@scut.edu.cn}



\author[1]{Delu Zeng}
\ead{dlzeng@scut.edu.cn}


\author[2]{John Paisley}

\ead{jwp2128@columbia.edu}

\cortext[cor1]{Corresponding author: Delu Zeng}
\begin{abstract}
Deep Gaussian process models typically employ discrete hierarchies, but recent advancements in differential Gaussian processes (DiffGPs) have extended these models to infinite depths. However, existing DiffGP approaches often overlook the uncertainty in kernel hyperparameters by treating them as fixed and time-invariant, which degrades the model’s predictive performance and neglects the posterior distribution. In this work, we introduce a fully Bayesian framework that models kernel hyperparameters as random variables and utilizes coupled stochastic differential equations (SDEs) to jointly learn their posterior distributions alongside those of inducing points. By incorporating the estimation uncertainty of hyperparameters, our method significantly enhances model flexibility and adaptability to complex dynamic systems. Furthermore, we employ a black-box adaptive SDE solver with a neural network to achieve realistic, time-varying posterior approximations, thereby improving the expressiveness of the variational posterior. Comprehensive experimental evaluations demonstrate that our approach outperforms traditional methods in terms of flexibility, accuracy, and other key performance metrics. This work not only provides a robust Bayesian extension to DiffGP models but also validates its effectiveness in handling intricate dynamic behaviors, thereby advancing the applicability of Gaussian process models in diverse real-world scenarios.
\end{abstract}


\end{frontmatter}



%
\section{Introduction}
\label{sec:intro}

Gaussian Process (GP) models \cite{rasmussen2003gaussian} are non-parametric Bayesian methods widely used for tasks such as regression, classification, and optimization. They provide a flexible way to model uncertainty by assuming a joint distribution over the input data and capturing correlations between data points through a kernel function. GPs have found extensive applications in forecasting, including predicting agricultural commodity prices \cite{jin2024forecasting, jin2024machine}, surrogate modeling \cite{marrel2024probabilistic}, Bayesian optimization \cite{wang2024pre}, and reinforcement learning \cite{zhao2023probabilistic}. However, despite their versatility, GP models face challenges when dealing with non-Gaussian distributions, complex distributions, time series data, and other difficult tasks.

To address this, deep Gaussian processes (DGPs) \cite{damianou2013deep}  have been introduced for more expressive representations. However, DGPs may also face learning issues if the individual GPs are not invertible \cite{duvenaud2014avoiding, dunlop2018deep}. One approach to address this challenges is via differential GPs (DiffGPs) \cite{hegde2018deep}, which model data evolution in continuous time using systems of stochastic differential equations. This approach enables the learning of continuous-time transformations of the data, providing a more intuitive means of capturing data dynamics compared to conventional techniques. DiffGPs warp inputs through differential fields to generalize discrete layers into a dynamic system. The intuition of ``warping'' the inputs over time, as derived from the original DiffGP paper, is an extension of the concept of deep Gaussian processes in continuous layers. Similar to the transition from ResNet \cite{he2016deep} to neural ODEs \cite{chen2018neural}, this method seeks to adaptively transform the input space over time, enabling the model to more effectively capture intricate patterns and dependencies within the data.

DiffGP methods \cite{hegde2018deep} often overlook the uncertainty in kernel hyperparameters, which is crucial for accurate model fitting and reliable uncertainty estimation. This uncertainty refers to the process of estimating and selecting these hyperparameters, which directly impacts the model’s predictive performance and the posterior distribution. The choice of covariance function in Gaussian processes significantly influences the shape of the posterior distribution and the uncertainty of predictions, highlighting the importance of selecting an appropriate covariance function \cite{lalchand2020approximate,lalchand2022sparse}.

In Gaussian process modeling, model selection involves not only choosing the covariance function but also estimating the hyperparameters. Traditionally, this process is performed by maximizing the marginal likelihood or its lower bound. However, the marginal likelihood is typically non-convex, especially when there are multiple local optima or when hyperparameters are weakly identifiable. This non-convexity makes optimization challenging, as gradient-based methods can get stuck in local minima, leading to overfitting and underestimating prediction uncertainty. Moreover, the sensitivity of gradient-based optimization to initial values further complicates the maximization of the marginal likelihood.

To overcome these issues, improved methods are needed to enhance the accuracy of hyperparameter estimation, which in turn will lead to better modeling of uncertainty and prevent overfitting. In this work, we propose a fully Bayesian approach to DiffGPs, treating kernel hyperparameters as random variables and using coupled stochastic differential equations (SDEs) to learn their posterior distribution and that of inducing points. This method effectively addresses the non-convexity of the marginal likelihood, providing robust parameter estimation. By incorporating uncertainty in hyperparameters through the posterior distribution, overfitting is avoided and generalization to new data improved. Bayesian methods offer a comprehensive understanding of parameter estimation by capturing inherent uncertainty, improving model reliability and adaptability.

Extending the Bayesian approach to include hyperparameters within a hierarchical framework increases the complexity of posterior computations, making the process more challenging. We introduce a novel methodology using SDEs to learn the posterior distribution of hyperparameters and inducing points, capturing their uncertainty effectively. By integrating Bayesian inference with SDEs, our approach enhances model adaptability and robustness in capturing system dynamics.

Our work introduces a novel network architecture that not only enhances the expressiveness of DiffGPs, but also integrates the uncertainty of kernel parameters and inducing point distributions into a fully Bayesian inference framework, incorporating their time-correlated posterior SDEs. By leveraging state-of-the-art SDE gradient estimators, such as those in \cite{li2020scalable, liu2020does, kong2020sde, tzen2019neural}, we demonstrate the effectiveness of approximate inference by maximizing our modified variational lower bound, significantly improving the scalability of gradient-based variational inference compared to previous studies. Computation of the output layer state is simplified using a black-box adaptive SDE solver, simplifying the modeling process and enhancing the efficiency of our proposed methodology.

Our approach offers two distinct advantages over previous DiffGP models:
1) It incorporates the uncertainty of inducing points and kernel hyperparameters within a fully Bayesian framework, which enhances flexibility and adaptability in capturing complex dynamics.
2) The SDE solver is responsible for describing how the posterior distributions of the kernel hyperparameters and inducing points evolve over time. By adopting an adaptive SDE approach, we achieve a comprehensive and realistic approximation of the posterior, ensuring that this approximation can effectively capture the dynamics of complex systems. This results in a more accurate posterior approximation that helps prevent overfitting.

Experimental evaluation demonstrates the improvement of our proposed method over traditional approaches in terms of flexibility, accuracy, and other metrics. Our contributions are outlined as follows:
\begin{itemize}
    \item We introduce a fully Bayesian approach that treats kernel hyperparameters as random variables and utilizes coupled stochastic differential equations (SDEs) to learn their posterior distribution and that of inducing points for DiffGPs. 
    \item By using an adaptive SDE method with a neural network as a black-box solver, we achieve a realistic posterior approximation that effectively captures time-varying dynamics and enhances the expressiveness of the variational posterior.
   \item  Experimental results show the advantage of our method over traditional approaches in terms of flexibility, accuracy, and other metrics.
\end{itemize}
The structure of this paper is organized as follows. Section \ref{sec:formatting} introduces the foundational models, including Gaussian Processes, Sparse Representations, and Continuous-time Gaussian Processes (DiffGPs). Section~\ref{section3} details our proposed method, called Fully Bayesian Differential Gaussian Processes (FB-Diff), along with the underlying methodologies and implementation details. Section \ref{section4} reviews related work and provides a comprehensive literature survey. Section \ref{section5} presents the experimental setup and results, demonstrating the effectiveness of our approach. Finally, Section \ref{section6} concludes the paper and discusses potential directions for future research.

\section{Model}
\label{sec:formatting}

\subsection{Gaussian Processes and Sparse Representation}
\label{2.1}
Gaussian processes (GPs) are powerful probabilistic models that define a distribution over functions. Given a set of input points $\mathbf{X} = [\mathbf{x}_1,\dots,\mathbf{x}_N]$, where each $x_i \in \mathbb{R}^{D}$, a GP specifies a joint Gaussian distribution over the corresponding function values $\mathbf{f} = [f(\mathbf{x}_1), \ldots, f(\mathbf{x}_N)]^\top \in \mathbb{R}^{N}$. The fundamental property of GPs is that the outputs at different points correlate based on their similarity, as measured by a kernel function $k(\mathbf{x}, \mathbf{x}')$ \cite{rasmussen2003gaussian}.

Traditionally, a zero-mean Gaussian process prior is defined on a function $f(\mathbf{x})$ over vector inputs $\mathbf{x} \in \mathbb{R}^D$:
\begin{equation}
f(\mathbf{x}) \sim \mathcal{GP}(0, k(\mathbf{x}, \mathbf{x}')),
\end{equation}
where $k(\mathbf{x}, \mathbf{x}')$ is the kernel function that captures the covariance between function values at different inputs.

To handle large datasets effectively, sparse Gaussian processes employ a small set of inducing or landmark variables \cite{titsias2009variational,paisley2010active,liang2015landmarking,xu2024sparse}. These inducing variables $\mathbf{u} = {u_1, \ldots, u_M}^\top \in \mathbb{R}^M$, where $M$ is much smaller than $N$, can be selected from the dataset or chosen independently. For example, Tran et al. (2021) \cite{tran2021sparse} propose utilizing neighbor information to guide the selection of inducing points, while Jafrasteh et al. (2022) \cite{jafrasteh2021input} suggest an input-dependent approach where inducing points are determined based on the input data characteristics. By conditioning the GP prior on these inducing variables $\mathbf{u}$ and their corresponding locations $\mathbf{Z} = \{\mathbf{z}_1, \ldots, \mathbf{z}_M\}^\top$, we can obtain posterior predictions at the data points.

The sparse GP posterior predictions, given the inducing variables $\mathbf{u}$ and their locations $\mathbf{Z}$, are given by
\begin{equation}
\begin{aligned}
\mathbf{f} \mid \mathbf{u}, \mathbf{Z} & \sim \mathcal{N}(\mathbf{Q}\mathbf{u}, \mathbf{K}_{\mathbf{XX}} - \mathbf{Q}\mathbf{K}_{\mathbf{ZZ}}\mathbf{Q}^\top) \\
\mathbf{u} & \sim \mathcal{N}(\mathbf{0}, \mathbf{K}_{\mathbf{ZZ}}).
\end{aligned}
\end{equation}
Here, \(\mathbf{Q} = \mathbf{K}_{\mathbf{XZ}}(\mathbf{K}_{\mathbf{ZZ}} + \sigma_n^2\mathbf{I})^{-1}\) represents the matrix of coefficients linking the inducing variables to the function values. In this expression, the subscript \(n\) denotes the index of the sample point,  \(\mathbf{K}_{\mathbf{ZZ}}\) is the pairwise kernel matrix for the inducing points, \(\mathbf{K}_{\mathbf{XX}}\) is the kernel matrix for the data points, \(\mathbf{K}_{\mathbf{XZ}}\) is the kernel matrix between the data points and the inducing points, and \(\sigma_n^2\) is the noise variance of the observations.

The main inference problems for Gaussian processes are related to the inducing points $\mathbf{Z}$, inducing variables $\mathbf{u}$, and the kernel hyperparameters $\mathbf{\lambda}$ found in, for example, the classic Gaussian kernel,
\begin{equation}
 k(\mathbf{x}, \mathbf{x}') = \sigma_f^2 \exp\left(-\frac{\|\mathbf{x} - \mathbf{x}'\|^2}{2l^2}\right).
\end{equation}
For model learning, variational inference (VI) \cite{hensman2015scalable} and Markov Chain Monte Carlo (MCMC) \cite{hensman2015mcmc} have been widely used. By incorporating sparse representation through inducing variables, we can effectively model and make predictions in large-scale datasets using Gaussian processes.

\subsection{Continuous-time Gaussian processes}
The continuous-time deep learning paradigm introduced by \cite{hegde2018deep} focuses on a model called DiffGP, which is a continuous-time deep Gaussian process model through infinite, infinitesimal differential compositions. In DiffGP models, the input data is transformed using a stochastic differential equation (SDE) flow to obtain transformed inputs $\mathbf{X}_T$, which are then used for model fitting after a predefined time $T$. The parameter \( T \) governs the flow's length and the system's capacity, akin to the number of layers in traditional deep neural networks or deep GPs.

In this framework, the inputs are redefined as temporal functions \(\mathbf{x}: \mathbb{T} \rightarrow \mathbb{R}^D\) over time, with state paths \(\mathbf{x}_t\) evolving over \(t \in \mathbb{T} = \mathbb{R}_{+}\). The observed inputs \(\mathbf{x}_{i, 0}\) represent the initial states. The goal is to classify or regress the final data points \(\mathbf{X}_T = \left(\mathbf{x}_{1, T}, \ldots, \mathbf{x}_{N, T}\right)^\top\), which represent the states after \(T\) time steps of an SDE flow, using a Gaussian process predictor. In the above notation, the first index represents the dimension of the sample points, indicating the number of sample particles. The second index represents the dimension of the discrete time points. The predictor, denoted as \( g\left(\mathbf{x}_T\right) \), is assumed to follow a Gaussian process prior with zero mean and covariance function \( K\left(\mathbf{x}_T, \mathbf{x}_T^{\prime}\right) \). When \( T = 0 \), the framework simplifies to a standard Gaussian process.

The prediction relies on the structure of the final dataset \(\mathbf{X}_T\), which is determined by the SDE flow \(d \mathbf{x}_t\) originating from the initial data \(\mathbf{X}\). We focus on SDE flows of the Ito type,
\begin{equation}
\label{eq2}
d \mathbf{x}_t=\boldsymbol{\mu}\left(\mathbf{x}_t\right) d t+\sqrt{\boldsymbol{\Sigma}\left(\mathbf{x}_t\right)} d W_t,
\end{equation}
where
$$
\begin{aligned}
& \boldsymbol{\mu}(\mathbf{x}_t)=\mathbf{K}_{\mathbf{x} \mathbf{Z}} \mathbf{K}_{\mathbf{Z}\mathbf{Z}}^{-1} \operatorname{vec}\left(\mathbf{U}\right), \\
& \boldsymbol{\Sigma}(\mathbf{x}_t)=\mathbf{K}_{\mathbf{x x}}-\mathbf{K}_{\mathbf{x} \mathbf{Z}} \mathbf{K}_{\mathbf{Z}\mathbf{Z}}^{-1} \mathbf{K}_{\mathbf{Z}\mathbf{x}} .
\end{aligned}
$$
The vector-valued Gaussian process is conditioned on the inducing variables \(\mathbf{U} = \left(\mathbf{u}_1, \ldots, \mathbf{u}_M\right)^\top\), which define the function values \(\mathbf{f}(\mathbf{z})\) at the inducing points \(\mathbf{Z} = \left(\mathbf{z}_1, \ldots, \mathbf{z}_M\right)\). In Equation (\ref{eq2}), \(\mathbf{U}\) has dimensions \(M \times D\), corresponding to a multi-output Gaussian process with output dimension \(D\). As a result, the dimension of \(\mathbf{K}_{\mathbf{Z} \mathbf{Z}}\) is \(MD \times MD\), representing a block matrix of kernels that capture the dependencies between the \(M\) inducing points and the \(D\) output dimensions.

The transformation process can be simulated using an SDE solver, such as Euler discretization \cite{yildiz2018learning},
\begin{equation}
    \mathbf{x}_{T} = \mathbf{x}_{0} + \int_{0}^{T}  \boldsymbol{\mu}(\mathbf{x}_t)\,dt + \int_{0}^{T} \sqrt{\Sigma\left(\mathbf{x}_t\right)} \,d W_t .
\end{equation}
The DiffGP model, which captures the time evolution and dynamics of data in a continuous manner, offers a more natural and flexible approach compared to traditional discretized approaches like Deep GPs \cite{damianou2013deep,salimbeni2017doubly}. However, the original DiffGP model has limitations. For example, it treats the kernel parameters as point estimates and does not consider their uncertainty using Bayesian inference methods. Additionally, it does not account for the temporal variability of the kernel parameters and inducing points, similar to how each layer in a deep GP has its own kernel parameters and inducing points. In the following section, we will introduce a new fully Bayesian framework that addresses these issues by utilizing recent advancements in SDE theory and its connection to variational inference.

\section{Fully Bayesian Variational Inference}
\label{section3}
\subsection{Modeling uncertainty in kernel hyperparameters and inducing points}
For Gaussian process (GP) regression models, the effectiveness of the model is closely tied to the selected kernel hyperparameters and inducing points, an inherently complex task. Finding optimal values for these parameters presents a challenge. Our proposed methodology employs a Bayesian framework to treat the uncertainty inherent in choosing kernel hyperparameters and inducing points.

In the generative process of the model, we assume prior distributions on the kernel hyperparameters $\boldsymbol{\lambda}_t$, inducing points $\mathbf{Z}_t$, and inducing variables $\mathbf{U}_t$. These various parameters control the behavior of the Gaussian process model and are crucial for determining the smoothness of the model. The inducing points $\mathbf{Z}_t$ are any set of points within the input space that can be optimized to approximate the function values at all possible input locations. The inducing variables $\mathbf{U}_t$ are the corresponding function values learned for the inducing points. 

Given these prior distributions, the model generates the observed data $\mathbf{X}$ and $\mathbf{y}$ by drawing samples from the Gaussian process defined by the kernel function with the hyperparameters $\boldsymbol{\lambda}_t$ and the inducing variables $\mathbf{U}_t$. For example, in the classic Gaussian kernel, $\boldsymbol{\lambda}=\{\sigma_f, l\}$. The posterior distribution of the hyperparameters, inducing points, and inducing variables can then be estimated based on the observed data using Bayesian inference techniques. This allows us to make predictions and infer the underlying structure of the data based on the Gaussian process model,
\begin{equation}
\label{forward}
\begin{aligned}
\text{Prior over hyperparameters}: & \quad \boldsymbol{\lambda}_t \sim p(\boldsymbol{\lambda}_t)
\\\text{Prior over inducing points}: & \quad \mathbf{Z}_t \sim p(\mathbf{Z}_t)
\\\text{Prior over inducing variables}: & \quad \mathbf{U}_t\sim \mathcal{GP}(0, K(\mathbf{Z}_t, \mathbf{Z}'_t))
\\\text{Model forward  SDE}: & \quad d \mathbf{x}_t=\boldsymbol{\mu}\left(\mathbf{x}_t\right) d t+\sqrt{\Sigma\left(\mathbf{x}_t\right)} d W_t
\\\text{Predictor Gaussian process}: & \quad\mathbf{g}\sim \mathcal{GP}(0,K\left(\mathbf{x}_T, \mathbf{x}_T^{\prime}\right))
\\\text{Data likelihood}: & \quad \mathbf{y} \mid \mathbf{g} \sim \mathcal{N}\left(\mathbf{g}, \sigma_n^2 \mathbb{I}\right)
\end{aligned}
\end{equation}

Inspired by the ANODEV2 method \cite{zhang2019anodev2}, which extends a complex ODE model for evolving neural network parameters, we view the hyperparameters $\boldsymbol{\lambda}_t$ and inducing points $\mathbf{Z}_t$ as prior processes that evolve over time. In contrast to traditional fully-GP models and previous works like DiffGPs, this approach departs from the fixed hyperparameters assumption and allows for a more dynamic modeling of the parameter evolution. Drawing parallels to evolutionary computing methods such as HyperNEAT \cite{stanley2009hypercube}, Compressed Weight Search  \cite{koutnik2010evolving}, and Hypernetworks \cite{DBLP:journals/corr/HaDL16}, which employ secondary networks to generate parameters for the main network, our approach extends this concept to parameter evolution in GP-SDE models. By adopting a Bayesian perspective, we aim to capture the time-varying distribution of hyperparameters, enhancing the adaptability of our model.

We can apply the concept of amortized variational inference, as introduced in the works of \cite{kim2018semi,agrawal2021amortized,lalchand2020approximate}, to optimize the factorized approximate posterior distribution $q(\boldsymbol{\lambda}_t,\mathbf{Z}_t,\mathbf{U}_t)=q(\boldsymbol{\lambda}_t)q(\mathbf{Z}_t)q(\mathbf{U}_t)$. This requirement is met by employing the mean-field assumption, where the posterior is approximated as a product of several independent factors. The goal is to minimize the Kullback-Leibler (KL) divergence between this approximate posterior and the true posterior distribution. This optimization objective is equivalent to maximizing the Evidence Lower Bound (ELBO), which serves as a lower bound on the marginal likelihood of the data under the model. By maximizing the ELBO, we can efficiently approximate the true posterior distribution and make accurate inference about the model's parameters and latent variables,
\begin{align}
\begin{aligned}
    \log p(\mathbf{y}) \ge & ~~ \mathbb{E}_{q\left( \boldsymbol{\lambda } \right) q\left( \mathbf{Z} \right) q\left( \mathbf{U} \right) p\left( \mathbf{x}|\boldsymbol{\lambda },\mathbf{Z},\mathbf{U} \right) p\left( \mathbf{g}\mid \mathbf{x} \right)}[\log p(\mathbf{y}\mid \mathbf{g})]
    \\
    &~~ -\mathrm{KL}\left( q\left( \boldsymbol{\lambda } \right) \parallel p\left( \boldsymbol{\lambda } \right) \right)\\
    &~~-\mathrm{KL}\left( q\left( \mathbf{Z} \right) \parallel p\left( \mathbf{Z} \right) \right) \\
    &~~-\mathrm{KL}\left( q\left( \mathbf{U} \right) \parallel p\left( \mathbf{U} \right) \right),
\end{aligned}
\end{align}
where $\boldsymbol{\lambda }$, $\mathbf{Z}$, and $\mathbf{U}$ represent the trajectories of $\boldsymbol{\lambda }_t$, $\mathbf{Z}_t$ and $\mathbf{U}_t$ for $t\in[0,T]$. We next demonstrate how to efficiently perform variational inference by optimizing the ELBO to estimate the prior hyperparameters and the parameters of a tractable approximate posterior.

\subsection{Approximate posterior through latent stochastic differential equations}
To perform posterior inference in our model, we leverage latent stochastic differential equations \cite{opper2019variational,li2020scalable,kidger2021neural,xu2022infinitely}. These SDEs provide a principled and flexible framework for modeling complex temporal dynamics. Specifically, we can represent both the prior and the approximate posterior of $\boldsymbol{\lambda }_t$ and $\mathbf{Z}_t$  using coupled SDEs in the following system of differential equations,
\begin{equation}
\label{approx}
\begin{aligned}
    &\begin{matrix}
	\begin{cases}
	d\boldsymbol{\lambda }_t=h_{\theta _{\lambda}}\left( \boldsymbol{\lambda }_t,t \right)dt +l_{\lambda}\left( \boldsymbol{\lambda }_t,t \right) dW_t                \quad(\mathrm{prior)}\\
	d\boldsymbol{\lambda }_t=h_{\phi _{\lambda}}\left( \boldsymbol{\lambda }_t,t \right)dt +l_{\lambda}\left( \boldsymbol{\lambda }_t,t \right) \mathrm{d}W_t  \quad(\mathrm{posterior~approx})\\
\end{cases}&		\,\,
	&		\,\,\\
\end{matrix}
\\&
\begin{matrix}
	\begin{cases}
	d\mathbf{Z}_t=h_{\theta _z}\left( \mathbf{Z}_t,t \right)dt +l_z\left( \mathbf{Z}_t,t \right) dW_t             \quad\,   (\mathrm{prior)}\\
	d\mathbf{Z}_t=h_{\phi _z}\left( \mathbf{Z}_t,t \right)dt +l_z\left( \mathbf{Z}_t,t \right) \mathrm{d}W_t      \quad\,  (\mathrm{posterior~approx})\\
\end{cases}&		\,\,
	&		\,\,\\
\end{matrix}
\end{aligned}
\end{equation}

The difference between the prior and posterior processes lies solely in their drift terms. Similar to classical Bayesian analysis, the parameters of the prior SDE can be set to simple, fixed values, such as constants or basic affine transformations, while the drift term of the posterior SDE is parameterized by a fully connected neural network. This allows us to compute the KL divergence between the prior and posterior SDEs using Girsanov's theorem \cite{van1976stochastic}. We apply this approach to both \(\mathbf{Z}\) and \(\boldsymbol{\lambda}\). Although we currently assume the drift term to be a simple function, it is also feasible to employ complex networks to learn the prior.

In the context of an Ornstein-Uhlenbeck (OU) prior stochastic differential equation (SDE), we specify fixed prior drift coefficients as $h_{\theta_\lambda}=-\boldsymbol{\lambda}_t$ and $h_{\theta_z}=-\mathbf{Z}_t$, along with fixed prior diffusion coefficients as $l_\lambda=\sigma_\lambda\mathbb{I}$ and $l_z=\sigma_z\mathbb{I}$. We choose the OU process because of its simplicity and well-studied properties. This process can be represented by the following SDE,
\begin{equation}
\begin{aligned}
    d\boldsymbol{\lambda }_t&=-\boldsymbol{\lambda }_tdt+\sigma _{\lambda}dW_t \quad (\mathrm{prior})\\
    d\mathbf{Z}_t&=- \mathbf{Z}_t dt +\sigma _{z} dW_t \quad(\mathrm{prior})
\end{aligned}
\end{equation}

For the approximate posterior processes of $\boldsymbol{\lambda }_t$ and $\mathbf{Z}_t$, we also employ an SDE representation, where $h_{\phi _{\lambda}}$ and $h_{\phi _z}$ are parameterized neural networks, and all drift and diffusion functions are Lipschitz continuous. Due to these drifts, the approximate posterior process will typically exhibit non-Gaussian, non-factorized marginals. It may seem restrictive to assume that the diffusion terms of the prior and posterior are identical in Equation (\ref{approx}). However, previous results from the Neural SDE literature demonstrate that any posterior can be approximated with arbitrary closeness using such a functional form given a sufficiently expressive drift process \cite{boue1998variational,tzen2019neural,li2020scalable,xu2022infinitely}. 

As a result, the Kullback-Leibler (KL) divergence between these distributions is finite and can be estimated by sampling paths from the posterior process \cite{opper2019variational,li2020scalable}. Using Girsanov's theorem, which states how probability measures change in stochastic processes under a change of drift, we can express the ELBO in a concise form,
\begin{equation}
\label{elbo}
\begin{aligned}
 \log p(\mathbf{y}) \geq & ~~ \mathrm{ELBO}=
\mathbb{E} _{q\left( \boldsymbol{\lambda } \right) q\left( \mathbf{Z} \right) q\left( \mathbf{U} \right) p\left( \mathbf{x}|\boldsymbol{\lambda },\mathbf{Z},\mathbf{U} \right) p\left( \mathbf{g}\mid \mathbf{x} \right)}[\log p(\mathbf{y}\mid \mathbf{g})]\\&~~
-{\frac{1}{2}}\int_0^T\mathbb{E} _{q\left( \boldsymbol{\lambda }_t \right)}[\left| u_{\lambda}\left( \boldsymbol{\lambda }_t,t \right) \right|^2]dt\\
&~~-{\frac{1}{2}}\int_0^T\mathbb{E}_{q\left( \mathbf{Z}_t \right)}[\left| u_z\left( \mathbf{Z}_t,t \right) \right|^2]dt-\mathrm{KL}\left( q\left( \mathbf{U} \right) \parallel p\left( \mathbf{U} \right) \right) 
\end{aligned}
\end{equation}
where
$$
\begin{aligned}
 &u_\lambda (\boldsymbol{\lambda }_t,t)=l_{\lambda}(\boldsymbol{\lambda }_t,t)^{-1}\left( h_{\theta _{\lambda}}(\boldsymbol{\lambda }_t,t)-h_{\phi _{\lambda}}(\boldsymbol{\lambda }_t,t) \right) ,
\\&
u_z(\mathbf{Z}_t,t)=l_z(\mathbf{Z}_t,t)^{-1}\left( h_{\theta _z}(\mathbf{Z}_t,t)-h_{\phi _z}(\mathbf{Z}_t,t) \right).
\end{aligned}
$$

The terms \( l_{\lambda}(\boldsymbol{\lambda}_t, t)^{-1} \) and \( l_z(\mathbf{Z}_t, t)^{-1} \) represent the left inverses, with the expectation being taken over the approximate posterior process defined by Equation (\ref{approx}).The functions $u_\lambda$ and $u_z$  need to fulfill the Novikov condition \cite{li2020scalable}, which ensures that the drift and diffusion terms of the SDE are well-defined and that the process remains consistent with the requirements of stochastic calculus.

All estimates of stochastic differential equation (SDE) paths and gradients are simulated and computed using state-of-the-art SDE solvers, as outlined in the work by \cite{li2020scalable}. Furthermore, to address large-scale data challenges efficiently, we can utilize mini-batch surrogates for likelihood optimization. This approach uses ideas from backpropagation introduced in \cite{rezende2014stochastic} and stochastic optimization techniques such as those discussed in \cite{amari1993backpropagation,hoffman2013stochastic,salimbeni2017doubly}. By employing mini-batch surrogates, we can optimize likelihood functions for models handling significant amounts of data while retaining computational efficiency,
\begin{equation}
  \log  p\left( \mathbf{y}|\mathbf{g} \right) =\sum_{i=1}^N{\log  p\left( y_i|\mathbf{g} \right)}\approx \frac{N}{B}\sum_{i=1}^B{\log  p\left( y_i|\mathbf{g} \right)}.
\end{equation}
For variational inference of the inducing variable $\mathbf{U}_t$, we follow the approach of \cite{hegde2018deep} by assuming that its posterior is a Gaussian distribution,
\begin{equation}
q\left( \mathbf{U}_t \right) =\mathcal{N} \left( \boldsymbol{m}_t, \boldsymbol{S}_t \right).
\end{equation}
What differentiates this paper from prior research is that $\boldsymbol{m}_t$ and $\boldsymbol{S}_t$ are time series while in the original DiffGP, this term is modeled as a constant. We aim to characterize the dynamical system of the inducing variables over time. Because the Kullback-Leibler divergence between two Gaussian distributions is analytical, the expression $\mathrm{KL}\left( q\left( \mathbf{U}_t \right) \parallel p\left( \mathbf{U}_t \right) \right)$ can be explicitly written as,
\begin{equation}
\label{eq13}
\begin{aligned}
&\mathrm{KL}\left( q\left( \mathbf{U} \right) \parallel p\left( \mathbf{U} \right) \right) =\int_0^T{\mathbb{E} _{q\left( \mathbf{U}_t,\boldsymbol{\lambda }_t,\mathbf{Z}_t \right)}\log \frac{q\left( \mathbf{U}_t \right)}{p\left( \mathbf{U}_t \right)}dt}\,\\&
~~~~~~~~~~={\frac{1}{2}}\int_0^T\mathbb{E} _{q} \left[ \begin{array}{c}
	\mathrm{trace} \left(K_{\mathbf{Z}_t\mathbf{Z}_t}^{-1}\boldsymbol{S}_t \right) +\boldsymbol{m}_{t}^{T}K_{\mathbf{Z}_t\mathbf{Z}_t}^{-1}\boldsymbol{m}_t+\ln \frac{| K_{\mathbf{Z}_t\mathbf{Z}_t}|}{|\boldsymbol{S}_t|}\\
\end{array} \right] dt
\end{aligned}
\end{equation}
In the second line, since  $\mathbf{U}_t$ has already been integrated out, there is no need for an expectation with respect to $\mathbf{U}_t$. However, the expectations with respect to  $\mathbf{Z}_t$ and $\boldsymbol{\lambda }_t$  are still necessary.

\subsection{Simulation and predictions}
 
By combining Equation (\ref{eq2}) with Equation (\ref{approx}), we have
 \begin{equation}
 \label{14}
d\left( \begin{array}{c}
	\mathbf{x}_t\\
	\boldsymbol{\lambda }_t\\
	\mathbf{Z}_t\\
\end{array} \right) =\left( \begin{array}{c}
	\boldsymbol{\mu }\left( \mathbf{x}_t \right)\\
	h_{\phi _{\lambda}}\left( \boldsymbol{\lambda }_t,t \right)\\
	h_{\phi _z}\left( \mathbf{Z}_t,t \right)\\
\end{array} \right) dt+\left( \begin{array}{c}
	\sqrt{\Sigma \left( \mathbf{x}_t \right)}\\
	l_{\lambda}\left( \boldsymbol{\lambda }_t,t \right)\\
	l_z\left( \mathbf{Z}_t,t \right)\\
\end{array} \right) dW_t.
 \end{equation}
By leveraging advanced SDE solvers for state trajectory approximation, we can perform stochastic gradient estimation to optimize Equation (\ref{elbo}). This technique enhances our understanding of the model's loss function and enables more efficient optimization. Integrating Bayesian methodologies for hyperparameters and inducing points with dynamic SDE posterior estimation leads to a more flexible and expressive posterior estimation for DiffGPs. This advancement improves the model's adaptability in capturing complex system dynamics and enhances predictive capabilities, providing a tool for robust model predictions. We present the algorithmic framework in Algorithm~\ref{alg:1}. As described in Section~\ref{2.1}, the GP sparse representation method significantly reduces the model complexity by employing the sparse inducing points approach \cite{titsias2009variational,paisley2010active,liang2015landmarking,xu2024sparse} in the DIffGP model. This decreases the computational complexity from $\mathcal{O}(M^3)$ to $\mathcal{O}(NM^2)$, where $M$ is the number of inducing points $\mathbf{Z}$, which is much smaller than $N$.

In summary, our method leverages the simulation of state trajectories as outlined in Equation (\ref{14}), allowing us to sample the entire model's predictive values. The integration of posterior parameter estimation through the SDE method provides a significant advantage over traditional approaches, enhancing the model’s robustness and improving uncertainty estimation. This approach improves the adaptability of the model to capture complex system dynamics as well as its predictive ability.

\begin{algorithm}[t]
\caption{FB-DiffGP Algorithm}
\label{alg:1}

\begin{algorithmic}
\REQUIRE $\mathbf{X}$: input data matrix; $\mathbf{Y}$: labels; $B$: mini-batch size; $E$: number of epochs; $\eta$: learning rate;
Parameter initialization for $\phi_\lambda$, $\phi_z$, $\boldsymbol{m}_t$, $\boldsymbol{S}_t$
\FOR{epoch = 1 to $E$}
    \STATE Shuffle the dataset $(\mathbf{X}, \mathbf{Y})$
    \FOR{each mini-batch $(\mathbf{X}_b, \mathbf{Y}_b)$ in $(\mathbf{X}, \mathbf{Y})$}
        \STATE \textbf{Compute ELBO} for the mini-batch by Equation (\ref{elbo}) and (\ref{eq13}),
       
        \STATE \textbf{Compute gradients} $\nabla_{\phi_\lambda} \mathcal{L}$, $\nabla_{\phi_z} \mathcal{L}$, $\nabla_{\boldsymbol{m}_t} \mathcal{L}$, $\nabla_{\boldsymbol{S}_t} \mathcal{L}$
        \STATE \textbf{Update parameters} using gradient ascent:
        \STATE \quad $\phi_\lambda\, \leftarrow \phi_\lambda\,\, + \eta \cdot \nabla_{\phi_\lambda} \mathcal{L}$
        \STATE \quad $\phi_z\,\, \leftarrow \phi_z\,\, + \eta \cdot \nabla_{\phi_z} \mathcal{L}$
        \STATE \quad $\boldsymbol{m}_t \leftarrow \boldsymbol{m}_t + \eta \cdot \nabla_{\boldsymbol{m}_t} \mathcal{L}$
        \STATE \quad $\boldsymbol{S}_t\,\, \leftarrow \boldsymbol{S}_t\,\, + \eta \cdot \nabla_{\boldsymbol{S}_t} \mathcal{L}$
    \ENDFOR
\ENDFOR
\RETURN $\phi_\lambda$, $\phi_z$, $\boldsymbol{m}_t$, $\boldsymbol{S}_t$
\end{algorithmic}
\end{algorithm}

\section{Related Work}
\label{section4}
\paragraph{Gaussian Processes (GPs)}  
GPs are a powerful class of non-parametric models widely used for regression, classification, and Bayesian optimization. They provide a probabilistic framework for modeling data by assuming that any finite subset of observations follows a multivariate Gaussian distribution. The flexibility of GPs allows them to capture complex patterns and uncertainty in data, making them ideal for applications such as predicting agricultural commodity prices \cite{jin2024forecasting, jin2024machine}, surrogate modeling \cite{marrel2024probabilistic}, Bayesian optimization \cite{wang2024pre}, and reinforcement learning \cite{zhao2023probabilistic}.  Despite their effectiveness, GPs suffer from a computational complexity of $\mathcal{O}(N^3)$, where $N$ is the number of training points. This high complexity arises from the need to compute and invert the $N \times N$ covariance matrix, which becomes prohibitive for large datasets.

\paragraph{Sparse GPs} Sparse Gaussian Processes are an extension of GPs that address scalability issues in modeling large datasets. Traditional GPs involve inverting the covariance matrix, which becomes computationally expensive as the size of the dataset increases. Sparse GPs alleviate this issue by introducing a smaller set of ``inducing points'' that approximate the latent function properties over the entire dataset. By assuming a joint distribution over the function values at the inducing points and the entire data set, Sparse GPs can approximate the true GP model while significantly reducing the computational complexity. Sparse GPs have gained popularity in various applications, including machine learning \cite{hensman2015scalable,sun2017location}, robotics \cite{schreiter2015sparse}, and computer vision \cite{tran2019calibrating}, where dealing with large datasets is common. Modeling and inference with Sparse GPs have evolved considerably over the last few years with key contributions in the direction of scalability to virtually any number of data points and generality within automatic differentiation frameworks \cite{matthews2017gpflow,krauth2016autogp, gardner2018gpytorch}. This has been possible thanks to the combination of stochastic variational inference techniques \cite{hoffman2013stochastic} with representations based on inducing variables \cite{titsias2009variational,lazaro2009inter,hensman2015scalable}. These advancements have now
made GPs attractive to a variety of applications and
likelihoods \cite{matthews2017gpflow,van2017convolutional,li2016review,corani2021time}. However, it is worth noting that Sparse GPs still require selecting the appropriate hyperparameters and inducing points, which can be challenging in some cases.

\paragraph{Fully Bayesian GPs} The key distinction between fully Bayesian GPs and traditional sparse Gaussian Processes is in their approach towards kernel hyperparameters. In traditional sparse Gaussian Processes, kernel hyperparameters are considered fixed or directly optimized model parameters. This means that during the modeling process, one needs to select a set of optimal hyperparameter values to fit the training data. While this approach can yield good results when the training data is abundant and of high quality, selecting appropriate hyperparameter values can become challenging in situations with scarce or nonlinear data. Fully Bayesian GPs treat kernel hyperparameters as random variables and introduce prior distributions to represent their uncertainty. This means that Fully Bayesian GPs no longer rely on fixed hyperparameter values but instead model the range of possible hyperparameter values and update their prior distributions based on the posterior distribution from the observed data. This approach enables the estimation of the true values of hyperparameters and their uncertainties using Bayesian inference, allowing for better adaptation to the data.

Fully Bayesian Gaussian processes have been used extensively by numerous researchers. In early studies, \cite{bernardo1998regression,williams1995gaussian} investigated the use of Hamiltonian Monte Carlo (HMC) methods to perform integration over covariance hyperparameters in the regression setting. \cite{barber1996gaussian} extended the application of HMC methods to the classification setting. They employed HMC for sampling in the hyperparameter space and utilized the Laplace approximation to compute the integral over function values. \cite{murray2010slice} focused on MCMC schemes to sample covariance hyperparameters in conjunction with latent function values, mainly mitigating the coupling effect through reparameterisation. \cite{hensman2015mcmc} considered joint sampling of inducing variables and hyperparameters from the optimal variational posterior distribution while \cite{bui2017streaming} considered inference schemes for fully Bayesian sparse GPs in
a streaming setting. The technique introduced by \cite{lalchand2020approximate} incorporates Variational Inference (VI) into Fully Bayesian GPs, approximating the posterior over hyperparameters with a factorized Gaussian distribution (mean-field approximation). More recently, \cite{rossi2021sparse} modified the generative model by adding a prior over the inducing inputs, and performed inference using SG-HMC over the joint  kernel hyperparameters $\boldsymbol{\lambda}$, inducing points $\mathbf{Z}$, inducing variables $\mathbf{U}$ space. Subsequently, \cite{lalchand2022sparse} extended this method to a doubly collapsed bound, which analytically selected the optimal distribution over the inducing points.

\paragraph{GPs with deep architectures} We also focus on the utilization of deep structures in GPs, specifically Deep Gaussian Processes (DGPs) and Differential Gaussian Processes (DiffGPs) consisting of discrete layers and continuous layers. DGP \cite{damianou2013deep} is a model composed of multiple layers of Gaussian processes. Each layer is a Gaussian process used to model the nonlinear relationships in intermediate layers. The output layer of the DGP provides the final prediction. In addition to traditional Bayesian inference methods, DGP introduces several new inference techniques. DSVI \cite{salimbeni2017doubly} introduced stochastic variational inference to handle large-scale data and learn the distribution of model parameters. SGHMC \cite{havasi2018inference} is a model that uses Hamiltonian Monte Carlo method for inference in deep Gaussian processes. This method incorporates stochastic gradients for learning, allowing for large-scale data processing during the inference process. IPVI \cite{yu2019implicit} constructed an approximate posterior by introducing Nash equilibrium. NOVI \cite{xu2023neural} is a method that uses neural networks and score-based approaches to approximate the complex posterior distribution in DGPs.  Unlike 
traditional DGP methods that focus on iterative function mappings, DiffGPs \cite{hegde2018deep}  utilized SDEs to characterize continuous-depth Gaussian processes. By transforming the GP modeling into a continuous-time framework and introducing SDEs to describe time evolution, DiffGPs can learn continuous-time transformations or flows of the data. Based on \cite{hegde2018deep}, \cite{solin2021scalable} derived direct approximations
to the  Fokker–Planck–Kolmogorov (FPK) equation in an assumed density Gaussian
form that avoids sampling-based inference in the latent space, which makes inference fast. Our work builds upon the model proposed by \cite{hegde2018deep} and combines it with the advantages of Fully Bayesian Gaussian processes. Additionally, we treat the hyperparameters as time-varying and utilize coupled neural SDEs for posterior inference of both the hyperparameters and inducing points. We summarize these methods in Table \ref{table1}.

\begin{table}[t!]
\resizebox{1.01\textwidth}{!}{
\begin{tabular}{ l r  cc cc cc cc }
\toprule
& Method name & Sparse &Layer & Time-vary & $\mathbf{U}$ & $\boldsymbol{\lambda}$ & $\mathbf{Z}$ & Inference & Ref. \\
\midrule
\multirow{1}{*}{GP}& GP  & $\times$
 & Single & -- & -- & Point & -- & Maximum likelihood  & \cite{rasmussen2003gaussian}  \\
\midrule
\multirow{2}{*}{Sparse GP}& FITC-SVGP  & $\checkmark$
 & Single & -- & Bayes & Point & Point & VI & \cite{titsias2009variational}  \\
& SVGP & $\checkmark$ & Single & -- & Bayes & Point & Point &SVI & \cite{hensman2015scalable}  \\
\midrule
\multirow{2}{*}{Bayesian GP}& SMCMC-GP  & $\checkmark$
 & Single & -- & Bayes & Bayes & Point & MCMC & \cite{hensman2015mcmc}  \\
& SSGP & $\checkmark$ & Single & -- & Bayes & Bayes & Point &  Streaming VI & \cite{bui2018partitioned}  \\
& Bayesian GPR & $\checkmark$ & Single & -- & Bayes & Bayes & Point &  VI/HMC & \cite{lalchand2020approximate}  \\
& BSGP & $\checkmark$ & Single & -- & Bayes & Bayes & Bayes &  SG-HMC & \cite{rossi2021sparse}  \\
&SGPR + HMC & $\checkmark$ & Single & -- & Bayes & Bayes & Point &  Collapsed VI& \cite{lalchand2022sparse}  \\
\cmidrule(lr){1-10}
\multirow{4}{*}{\shortstack{Deep GP}} & DGP & $\times$ & Discrete & -- & -- & Point & -- & VI & \cite{damianou2013deep}  \\
& DSVI-DGP  & $\checkmark$ & Discrete & -- & Bayes  &Point & Point & DSVI& \cite{salimbeni2017doubly} \\
& SGHMC-DGP & $\checkmark$ & Discrete & -- &Bayes  &Point &Point  &  SGHMC & \cite{havasi2018inference}  \\
& IPVI-DGP & $\checkmark$  & Discrete& -- & Bayes&Point & Point & IPVI& \cite{yu2019implicit}  \\
& NOVI-DGP      & $\checkmark$ & Discrete& --  & Bayes  & Point&Point &  NOVI & \cite{xu2023neural}  \\
\cmidrule(lr){1-10}
\multirow{2}{*}{\shortstack{ Cont-time GP  }}
& DiffGP &$\checkmark$ & Continuous & $\times$ & Bayes & Point &Point & SVI & \cite{hegde2018deep} \\ 
& Match FPK & $\checkmark$ & Continuous &$\times$ & Bayes & Point & Point &  Assumed density & \cite{solin2021scalable} \\
\cmidrule(lr){1-10}
\multirow{1}{*}{\shortstack{Ours}}
& FB-DiffGP &$\checkmark$ & Continuous & $\checkmark$ & Bayes & Bayes & Bayes & SVI + neural  SDE & -- \\ 

\bottomrule
\end{tabular}
}
\caption{We summarize the existing literature on sparse GPs, deep GPs, and continuous-time GPs, focusing on the handling methods for kernel hyperparameters $\boldsymbol{\lambda}$, inducing points $\mathbf{Z}$, inducing variables $\mathbf{U}$, and the associated inference techniques, whether through point estimation or Bayesian estimation. From this table, it is evident that our approach incorporates the idea of continuous layers and employs time-varying Bayesian posterior inference, significantly enhancing the model's flexibility and robustness.}
\label{table1}
\end{table}

\paragraph{SDE solvers} SDE solvers are numerical methods for approximating the state trajectories of the system over time by discretizing the SDEs into steps and iteratively updating the state variables. They are widely used in scientific and engineering fields, especially in computing Neural SDEs \cite{liu2019neural,kong2020sde,liu2020does}. In Neural SDEs, the drift and diffusion terms of the SDE are parameterized using neural networks. This allows for more expressive modeling of the dynamics compared to traditional SDEs. SDE solvers are vital for training and inference in Neural SDE models. By approximating state trajectories, SDE solvers enable the calculation of gradients \cite{li2020scalable,kidger2021efficient}, which is necessary for optimizing neural network parameters using techniques like stochastic gradient descent. Moreover, SDE solvers are also used in the simulation and generation of data from the learned Neural SDE models. These solvers enable the generation of synthetic data that captures the complex dynamics of the modeled systems, making it possible to analyze and explore the behavior of the neural system under different conditions. The use of SDE solvers in Neural SDEs has shown remarkable results in various machine learning applications including time series forecasting \cite{yang2023neural}, generative modeling \cite{ho2020denoising}, uncertainty quantification, and reinforcement learning \cite{chen2019incremental}. 

\paragraph{Relationship between Generative Artificial Intelligence Approaches}

Neural Stochastic Differential Equation (SDE) methods, particularly score-based diffusion SDE models, have been extensively applied in the Generative Artificial Intelligence (GAI) field for tasks such as image synthesis \cite{song2020score, ma2022accelerating}, 3D generation \cite{hong2023debiasing}, and audio creation \cite{rouard2021crash}. These approaches leverage SDE solvers to generate realistic and high-quality data by modeling complex data distributions through diffusion processes. Although both our method and GAI approaches utilize black-box SDE solvers, the underlying objectives and theoretical foundations differ significantly. 

Our approach is grounded in SDE theory, specifically leveraging Girsanov's theorem and Variational Inference (VI) methods to address the hyperparameter uncertainty inherent in traditional continuous-time Gaussian processes. This allows for a more principled and theoretically sound handling of uncertainty in model parameters, enhancing the robustness and reliability of the predictive models. In contrast, generative AI primarily employs SDE solvers to generate data that closely resembles real-world samples, focusing on the quality and realism of the generated outputs rather than on resolving parameter uncertainties. Consequently, although both methodologies utilize similar computational tools, their goals and theoretical underpinnings cater to distinct aspects of machine learning and data modeling. This distinction underscores the versatility of SDE-based approaches in addressing a wide range of challenges across different domains.

\section{Experiments}
\label{section5}
We evaluate the performance of our approximate inference method, FB-DiffGP, on UCI datasets for regression and classification, comparing it against state-of-the-art techniques: SVGP \cite{hensman2015scalable}, DGP \cite{salimbeni2017doubly}, and DiffGP \cite{hegde2018deep}. The reasons for selecting these baseline methods are as follows: DiffGP is the primary baseline for our approach, making it the main focus of our comparison because it provides an approximate inference method for continuous deep Gaussian processes. DGP \cite{salimbeni2017doubly}, another important work on deep Gaussian processes, constructs a deep network structure to effectively capture the complexity of data. As FB-DiffGP shares similarities in terms of model architecture, it serves as a meaningful point of comparison. SVGP \cite{hensman2015scalable} is a widely used scalable sparse variational inference method in Gaussian process regression that performs well on large-scale datasets.

The number of inducing points $M$ is manually selected to balance accuracy and computation time using cross validation. We optimize all parameters jointly using the evidence lower bound and employ stochastic optimization with mini-batches and the Adam optimizer. Numerical solutions of SDEs are obtained using the Euler-Maruyama solver with 20 time steps.   The number of steps changes if the time interval $T$ is increased. For larger $T$, we would typically need to increase the number of time steps to maintain the same level of accuracy in the discretization of the SDE. This ensures that the solution remains precise and the numerical methods used are effective for longer time intervals. Other solvers, such as higher-order or adaptive methods, can also be easily implemented using Python toolkits.

Our model primarily handles classification and regression tasks, employing a simple network architecture with a few fully connected layers. The mini-batch size chosen is $10,000$ and the learning rate is set to $10^{-2}$. Our implementation utilizes GPyTorch  \cite{gardner2018gpytorch}, a Gaussian processes framework based on PyTorch. For comparison, we used a simple two-layer binarized neural network (BNN). The network includes a hidden layer with a defined number of neurons, followed by an output layer. The DNN used for comparison is the same as described in \cite{baldi2014searching}. This network architecture includes several layers, with details provided in that paper. Our code is available at \url{https://github.com/xujianscut/FB-DIFFGP}.

\subsection{Description of Datasets}
Our experiments encompass both unsupervised dimensionality reduction clustering tasks and supervised Bayesian classification and regression tasks. 

For unsupervised learning, presented in Section~\ref{us}, we utilize the multi-phase Oilflow dataset \cite{bishop1993analysis}. This dataset consists of measurements from multi-phase oil flow processes, capturing the dynamic behavior of oil, water, and gas interactions. It is important for assessing clustering algorithms in industrial scenarios with complex flow patterns. 

For supervised learning, presented in Sections~\ref{rb} and~\ref{cb}, we employ eight benchmark regression datasets from the UCI Machine Learning Repository \cite{asuncion2007uci}. These datasets include:
\begin{itemize}
    \item \textbf{Yacht Hydrodynamics}: Predicts the hydrodynamic performance of sailing yachts based on various design parameters.
    \item \textbf{Boston Housing}: Estimates median house prices in Boston suburbs using features such as crime rate, number of rooms, and accessibility to highways.
    \item \textbf{Energy Efficiency}: Models the heating and cooling load requirements of buildings based on architectural and environmental factors.
    \item \textbf{Concrete Compressive Strength}: Predicts the compressive strength of concrete using ingredients like cement, slag, and age.
    \item \textbf{Power Plant}: Forecasts the electrical power output of a power plant based on ambient variables and operating conditions.
    \item \textbf{Elevators}: Estimates the shaft size required for elevators in buildings based on building characteristics and elevator specifications.
    \item \textbf{Protein Tertiary Structure}: Predicts the spatial coordinates of protein structures from amino acid sequences.
    \item \textbf{Year Prediction MSD}: Forecasts the release year of songs based on audio features extracted from the Million Song Dataset.
\end{itemize}
These regression datasets vary in size, with the number of data points ranging from 308 (Yacht Hydrodynamics) to 515,345 (Year Prediction MSD). They originate from diverse real-world applications, demonstrating the scalability and applicability of our proposed method to practical regression problems across different domains.

For classification, we utilize the SUSY \cite{susy_279} and HIGGS \cite{higgs_280} datasets, which are large-scale real-world datasets containing 5,500,000 and 11,000,000 samples, respectively. 
\begin{itemize}
    \item \textbf{SUSY}: This dataset involves the classification of events in high-energy physics experiments to distinguish between signal events (indicative of supersymmetry) and background noise. It is essential for evaluating the performance of classification algorithms in handling high-dimensional and large-scale data typical in particle physics.
    \item \textbf{HIGGS}: Designed for the classification of particle collision events to identify the production of Higgs bosons, this dataset is crucial for assessing the efficiency and operational effectiveness of classification methods in processing massive datasets prevalent in scientific research and industry applications.
\end{itemize}
Both SUSY and HIGGS datasets are derived from real-world high-energy physics experiments. They enable us to evaluate the scalability of our proposed approach. The diverse range of datasets employed in our experiments, encompassing both regression and classification tasks from various real-world domains, highlights the robustness and versatility of our FB-Diff approach.

\subsection{ Unsupervised learning}
\label{us}
\begin{figure}[t]
    \centering
    \begin{subfigure}{0.323\textwidth}
        \centering
        \includegraphics[width=\textwidth,height=2in]{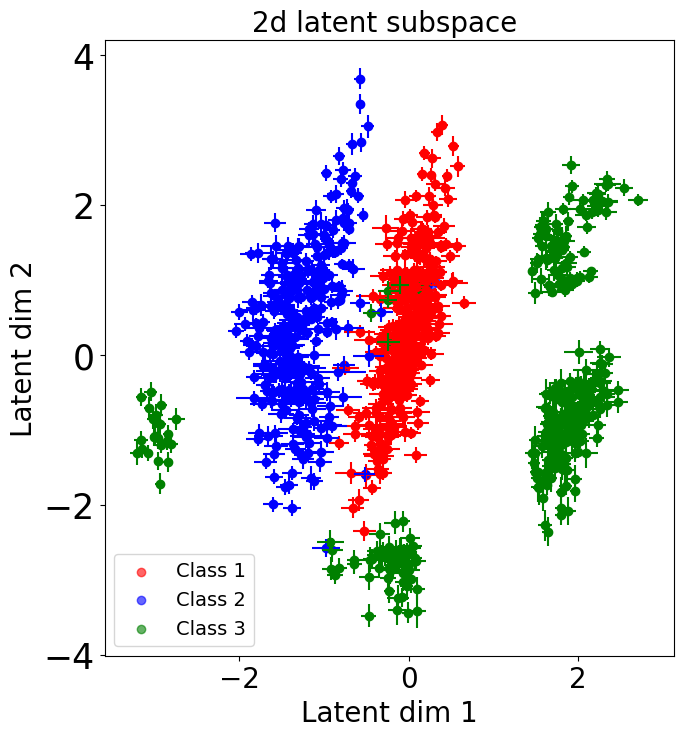}
        \caption{FB-DiffGP (Ours)}
    \end{subfigure}
    \begin{subfigure}{0.323\textwidth}
        \centering
        \includegraphics[width=\textwidth,height=2in]{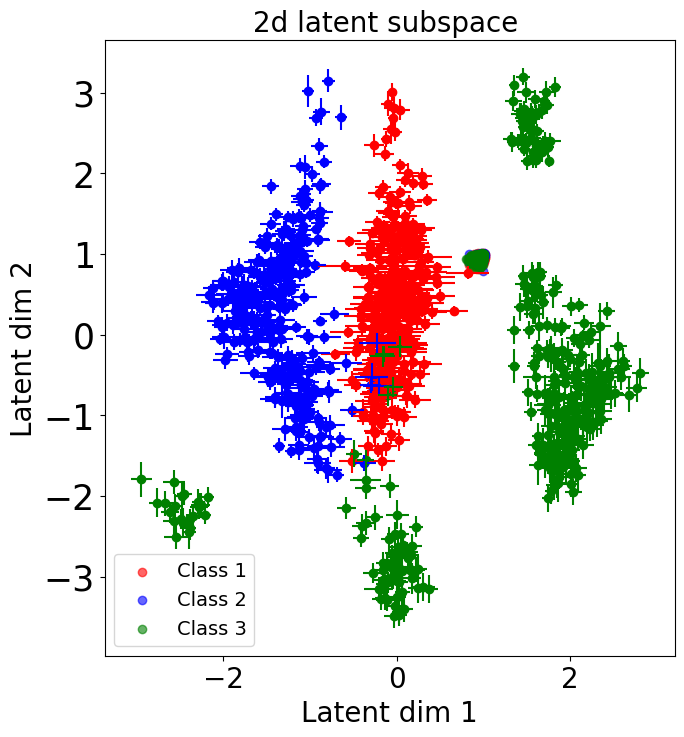}
        \caption{DiffGP}
    \end{subfigure}
    \begin{subfigure}{0.323\textwidth}
        \centering
        \includegraphics[width=\textwidth,height=2in]{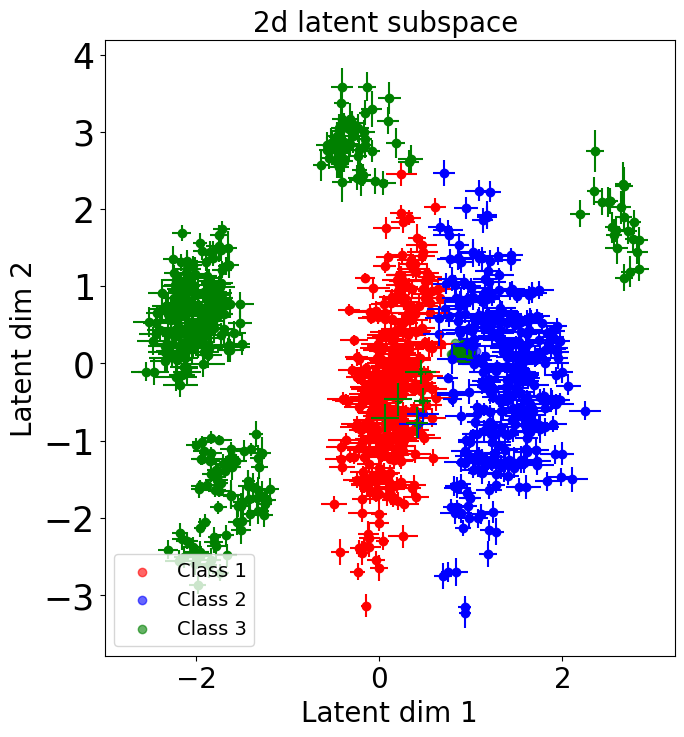}
        \caption{Standard GPLVM}
    \end{subfigure}
    \caption{Unsupervised dimensionality reduction on the ``Oilflow'' toy data set.}
    \label{f1}
\end{figure}

We applied our model to a dimensionality reduction task using the Bayesian Gaussian Process Latent Variable Model (GPLVM) \cite{titsias2010bayesian} for data reconstruction. Our toy dataset is the multi-phase Oilflow data \cite{bishop1993analysis}, comprising 1,000 data points in 12 dimensions, categorized into three classes representing different phases of oil flow in a pipeline. We reduced the data dimensionality to 10 while retaining as much information as possible.  As the training is unsupervised, the ground-truth labels were not used during training. We present the reconstruction error and mean squared error (MSE) along with $\pm$2 standard errors from ten optimization runs. The 2D projections of the latent space for the Oilflow data clearly show that our model effectively reveals the class structure. To emphasize the strengths of our model, we present the results of the 2D latent space for three models in Figure \ref{f1}. As shown in Figure \ref{f1}, the reconstructed data points for the three classes are more distinctly separated in our model, resulting in improved and more intuitive clustering. From Table \ref{tab:un}, we observe that our proposed FB-DiffGP outperforms both the Standard GPLVM and DiffGP methods, yielding lower reconstruction loss and improved uncertainty estimation.

\begin{table}[t!]
\centering
\begin{tabular}{|c|c|c|}
\hline
  Method & MSE & NLL  \\
\hline
 Standard GPLVM & 2.45 $\pm$ 0.05 & -12.42 $\pm$ 0.07 \\
\hline
DiffGP & 1.87 $\pm$ 0.04 &-14.41 $\pm$ 0.04\\
\hline
FB-DiffGP (Ours) & 1.65 $\pm$ 0.03 &-16.51 $\pm$ 0.05\\
\hline
\end{tabular}
\caption{ MSE and NLL for our FB-DiffGP compared to the baseline and standard GPLVM on toy ``Oilflow'' dataset of 1000 points in 12 dimensions.}
\label{tab:un}
\end{table}

\begin{table}[thp!]
\centering
\resizebox{.9\textwidth}{!}{
\begin{tabular}{ l r  c c c c }
\toprule
& & Yacht & Boston & Energy & Concrete     \\
\cmidrule(lr){3-6}
& $N\,|\,D$ &308 $|$ 6& 506 $|$ 13 & 768 $|$ 8 & 1,030 $|$ 8  \\
\midrule
Linear & & 0.68 $\pm$ 0.05 & 4.24 $\pm$ 0.16  & 2.88 $\pm$ 0.05 & 10.54 $\pm$ 0.13   \\
\cmidrule(lr){1-6}
BNN & $L=2$& 0.47 $\pm$ 0.04 &3.01 $\pm$ 0.18 & 1.80 $\pm$ 0.05 & 5.67 $\pm$ 0.09    \\
\cmidrule(lr){1-6}
\multirow{2}{*}{Sparse GP}& $M=100$  &0.45 $\pm$ 0.04 & 2.87 $\pm$ 0.15 & 0.78 $\pm$ 0.02 & 5.97 $\pm$ 0.11  \\
& $M=500$  &0.44 $\pm$ 0.04 & 2.73 $\pm$ 0.12 & 0.47 $\pm$ 0.02 & 5.53 $\pm$ 0.12  \\
\cmidrule(lr){1-6}
\multirow{4}{*}{\shortstack{Deep GP \\ $M=100$}} & $L=2$ &0.45 $\pm$ 0.03 & 2.90 $\pm$ 0.17 & 0.47 $\pm$ 0.01 & 5.61 $\pm$ 0.10\\
& $L=3$& 0.45 $\pm$ 0.03 & 2.93 $\pm$ 0.16 & 0.48 $\pm$ 0.01 & 5.64 $\pm$ 0.10  \\
& $L=4$      & 0.44 $\pm$ 0.03 &2.90 $\pm$ 0.15 & 0.48 $\pm$ 0.01 & 5.68 $\pm$ 0.10  \\
& $L=5$      &0.42 $\pm$ 0.03 & 2.92 $\pm$ 0.17 & 0.47 $\pm$ 0.01 & 5.65 $\pm$ 0.10\\
\cmidrule(lr){1-6}
\multirow{5}{*}{\shortstack{DiffGP \\ $M=100$}}& $T= 1.0$ &0.45 $\pm$ 0.04 &2.80 $\pm$ 0.13 & 0.49 $\pm$ 0.02 & 5.32 $\pm$ 0.10 \\ 
& $T= 2.0$ &0.43 $\pm$ 0.04 & 2.68 $\pm$ 0.10 & 0.48 $\pm$ 0.02 & 4.96 $\pm$ 0.09\\
& $T= 3.0$ &0.43 $\pm$ 0.03 & 2.69 $\pm$ 0.14 & 0.47 $\pm$ 0.02 & 4.76 $\pm$ 0.12   \\
& $T= 4.0$ &0.42 $\pm$ 0.03 & 2.67 $\pm$ 0.13 & 0.49 $\pm$ 0.02 & 4.65 $\pm$ 0.12  \\
& $T= 5.0$ &0.40 $\pm$ 0.04 & 2.58 $\pm$ 0.12 & 0.50 $\pm$ 0.02 & 4.56 $\pm$ 0.12 \\
\cmidrule(lr){1-6}
\multirow{5}{*}{\shortstack{FB-DiffGP (ours) \\ $M=100$}}
& $T= 1.0$ &\textbf{0.43} $\pm$ 0.04 &\textbf{2.63} $\pm$ 0.10 & \textbf{0.42} $\pm$ 0.01 & \textbf{4.75} $\pm$ 0.12  \\ 
& $T= 2.0$ &\textbf{0.41} $\pm$ 0.03 & \textbf{2.49} $\pm$ 0.10 & \textbf{0.41} $\pm$ 0.02 & \textbf{4.33} $\pm$ 0.11 \\
& $T= 3.0$ &\textbf{0.41} $\pm$ 0.03 & \textbf{2.47} $\pm$ 0.11 & \textbf{0.39} $\pm$ 0.01 & \textbf{4.22} $\pm$ 0.12  \\
& $T= 4.0$ &\textbf{0.40} $\pm$ 0.02 & \textbf{2.45} $\pm$ 0.09 & \textbf{0.37} $\pm$ 0.02 & \textbf{4.07} $\pm$ 0.11  \\
& $T= 5.0$ &\textbf{0.38} $\pm$ 0.04 & \textbf{2.39} $\pm$ 0.10 & \textbf{0.38} $\pm$ 0.01 & \textbf{4.01} $\pm$ 0.11  \\
\bottomrule
\toprule
& &  Power  & Elevators & Protein & Year  \\
\cmidrule(lr){3-6}
& $N\,|\,D$ & 9,568 $|$ 4   & 16,599 $|$ 18 & 45,730 $|$ 9 &515,345 $|$ 90  \\
\midrule
Linear & &  4.51 $\pm$ 0.03 & 5.08 $\pm$ 0.03 & 5.21 $\pm$ 0.02 & 6.35 $\pm$ 0.05  \\
\cmidrule(lr){1-6}
BNN & $L=2$&  4.12 $\pm$ 0.03 &4.57 $\pm$ 0.03  & 4.73 $\pm$ 0.01 &5.78 $\pm$ 0.05  \\
\cmidrule(lr){1-6}
\multirow{2}{*}{Sparse GP}& $M=100$  & 3.91 $\pm$ 0.03 & 4.47 $\pm$ 0.03 & 4.43 $\pm$ 0.03 &5.59 $\pm$ 0.06  \\
& $M=500$  & 3.79 $\pm$ 0.03  &4.32 $\pm$ 0.03 & 4.10 $\pm$ 0.03 &5.33 $\pm$ 0.04  \\
\cmidrule(lr){1-6}
\multirow{4}{*}{\shortstack{Deep GP \\ $M=100$}} & $L=2$ & 3.79 $\pm$ 0.03 & 4.35 $\pm$ 0.04  & 4.00 $\pm$ 0.03 &5.43 $\pm$ 0.04 \\
& $L=3$&  3.73 $\pm$ 0.04 &4.34 $\pm$ 0.03  & 3.81 $\pm$ 0.04 &5.38 $\pm$ 0.04  \\
& $L=4$      &   3.71 $\pm$ 0.04 & 4.32 $\pm$ 0.03 & 3.74 $\pm$ 0.04 &5.25 $\pm$ 0.03  \\
& $L=5$      &  3.68 $\pm$ 0.03 &4.30 $\pm$ 0.03  & \textbf{3.72} $\pm$ 0.04 &\textbf{5.23} $\pm$ 0.03 \\
\cmidrule(lr){1-6}
\multirow{5}{*}{\shortstack{DiffGP \\ $M=100$}}& $T= 1.0$ & 3.76 $\pm$ 0.03 &4.38 $\pm$ 0.03  & 4.04 $\pm$ 0.04 &5.45 $\pm$ 0.04 \\ 
& $T= 2.0$ &    3.72 $\pm$ 0.03 &4.33 $\pm$ 0.03  & 4.00 $\pm$ 0.04 &5.41 $\pm$ 0.03 \\
& $T= 3.0$ &    3.68 $\pm$ 0.03 &4.32 $\pm$ 0.03  & 3.92 $\pm$ 0.04 &5.37 $\pm$ 0.03  \\
& $T= 4.0$ &  3.66 $\pm$ 0.03 &4.30 $\pm$ 0.03  & 3.89 $\pm$ 0.04 &5.33 $\pm$ 0.03 \\
& $T= 5.0$ &   3.65 $\pm$ 0.03 &4.30 $\pm$ 0.02  & 3.87 $\pm$ 0.04 &5.30 $\pm$ 0.03  \\
\cmidrule(lr){1-6}
\multirow{5}{*}{\shortstack{FB-DiffGP (ours) \\ $M=100$}}
& $T= 1.0$ &    \textbf{3.64} $\pm$ 0.03 & \textbf{4.32} $\pm$ 0.03 & \textbf{3.94} $\pm$ 0.03 &\textbf{5.40} $\pm$ 0.04 \\ 
& $T= 2.0$ &     \textbf{3.61} $\pm$ 0.03 & \textbf{4.30} $\pm$ 0.03 & \textbf{3.88} $\pm$ 0.03 &\textbf{5.36} $\pm$ 0.03\\
& $T= 3.0$ &    \textbf{3.58} $\pm$ 0.03 & \textbf{4.28} $\pm$ 0.04 & \textbf{3.85} $\pm$ 0.03 &\textbf{5.33} $\pm$ 0.03  \\
& $T= 4.0$ & \textbf{3.54} $\pm$ 0.03 & \textbf{4.25} $\pm$ 0.03 & \textbf{3.81} $\pm$ 0.03 &\textbf{5.28} $\pm$ 0.03 \\
& $T= 5.0$ &  \textbf{3.53} $\pm$ 0.03 &\textbf{4.25} $\pm$ 0.02 & \textbf{3.79} $\pm$ 0.03 &\textbf{5.24} $\pm$ 0.03  \\
\bottomrule
\end{tabular}
}
\caption{The test RMSE values on 8 benchmark datasets using 90\%/10\% random training and test splits with 20 repetitions. }
\label{rmse}
\end{table}

\begin{table}[thp!]
\centering
\resizebox{.90\textwidth}{!}{
\begin{tabular}{ l r  cc cc }
\toprule
& & Yacht & Boston & Energy & Concrete   \\
\cmidrule(lr){3-6}
& $N\,|\,D$ &308 $|$ 6& 506 $|$ 13 & 768 $|$ 8 & 1,030 $|$ 8    \\
\midrule
Linear &&-0.72 $\pm$ 0.03 & -2.89 $\pm$ 0.03 &-2.48 $\pm$ 0.02 & -3.78 $\pm$ 0.01   \\
\cmidrule(lr){1-6}
BNN & $L=2$&-0.64 $\pm$ 0.06 & -2.57 $\pm$ 0.09 & -2.04 $\pm$ 0.02 & -3.16 $\pm$ 0.02 \\
\cmidrule(lr){1-6}
\multirow{2}{*}{Sparse GP}& $M=100$  &-0.61 $\pm$ 0.04 & -2.47 $\pm$ 0.05 & -1.29 $\pm$ 0.02 & -3.18 $\pm$ 0.02
\\
& $M=500$  &-0.57 $\pm$ 0.03 & -2.40 $\pm$ 0.07 &  - 0.93 $\pm$ 0.01 &-3.09 $\pm$ 0.02  \\
\cmidrule(lr){1-6}
\multirow{4}{*}{\shortstack{Deep GP \\ $M=100$}} & $L=2$      &-0.61 $\pm$ 0.05 &-2.47 $\pm$ 0.05 & -0.73 $\pm$ 0.02 & -3.12 $\pm$ 0.01  \\
& $L=3$&-0.58 $\pm$ 0.04      & -2.49 $\pm$ 0.05 & -0.75 $\pm$ 0.02 & -3.13 $\pm$ 0.01   \\
& $L=4$      &-0.60 $\pm$ 0.04 &-2.48 $\pm$ 0.05 & -0.76 $\pm$ 0.02 & -3.14 $\pm$ 0.01   \\
& $L=5$      &-0.58 $\pm$ 0.03 &-2.49 $\pm$ 0.05 &-0.74 $\pm$ 0.02 & -3.13 $\pm$ 0.01    \\
\cmidrule(lr){1-6}
\multirow{5}{*}{\shortstack{DiffGP \\ $M=100$}}
&$T= 1.0$ &-0.57 $\pm$ 0.04 &-2.36 $\pm$ 0.04 & -0.65 $\pm$ 0.03 & -3.05 $\pm$ 0.02 \\ 
&$T= 2.0$ &-0.55 $\pm$ 0.03 & -2.32 $\pm$ 0.04 &-0.63 $\pm$ 0.03 & -2.96 $\pm$ 0.02\\ 
&$T= 3.0$ &-0.53 $\pm$ 0.03 &-2.31 $\pm$ 0.05 &-0.63 $\pm$ 0.03 & -2.93 $\pm$ 0.04\\ 
&$T= 4.0$ &-0.56 $\pm$ 0.05 &-2.33 $\pm$ 0.06 &-0.65 $\pm$ 0.03 & -2.91 $\pm$ 0.04\\ 
&$T= 5.0$ &-0.54 $\pm$ 0.03 &-2.30 $\pm$ 0.05 &-0.66 $\pm$ 0.03 & -2.90 $\pm$ 0.05\\ 
\cmidrule(lr){1-6}
\multirow{5}{*}{\shortstack{FB-DiffGP (ours) \\ $M=100$}}
&$T= 1.0$ &\textbf{-0.51} $\pm$ 0.03 &\textbf{-2.28} $\pm$ 0.04 &\textbf{-0.62} $\pm$ 0.03 &  \textbf{-2.75} $\pm$ 0.03\\ 
&$T= 2.0$ & \textbf{-0.49} $\pm$ 0.03 &\textbf{-2.26} $\pm$ 0.05 &\textbf{-0.59} $\pm$ 0.03 &  \textbf{-2.72} $\pm$ 0.03\\ 
&$T= 3.0$ &\textbf{-0.47} $\pm$ 0.03 &\textbf{-2.23} $\pm$ 0.04 &\textbf{-0.56} $\pm$ 0.04 &  \textbf{-2.63} $\pm$ 0.03\\ 
&$T= 4.0$ &\textbf{-0.44} $\pm$ 0.04 &\textbf{-2.21} $\pm$ 0.05 &\textbf{-0.54} $\pm$ 0.03 &  \textbf{-2.63} $\pm$ 0.04\\ 
&$T= 5.0$ &\textbf{-0.43} $\pm$ 0.03 &\textbf{-2.19} $\pm$ 0.04 & \textbf{-0.52} $\pm$ 0.02 &-\textbf{2.60} $\pm$ 0.03\\ 
\bottomrule
\toprule
& &  Power  & Elevators & Protein & Year  \\
\cmidrule(lr){3-6}
& $N\,|\,D$ & 9,568 $|$ 4   & 16,599 $|$ 18 & 45,730 $|$ 9 &515,345 $|$ 90  \\
\midrule
Linear &&-2.93 $\pm$ 0.01 &-2.76 $\pm$ 0.03  & -3.07 $\pm$ 0.00 &-6.32 $\pm$ 0.03  \\
\cmidrule(lr){1-6}
BNN & $L=2$&-2.84 $\pm$ 0.01 &-2.45 $\pm$ 0.05 & -2.97 $\pm$ 0.00 &-5.62 $\pm$ 0.03  \\
\cmidrule(lr){1-6}
\multirow{2}{*}{Sparse GP}& $M=100$  & -2.75 $\pm$ 0.01 &-2.41 $\pm$ 0.04  & -2.91 $\pm$ 0.00 &-5.46 $\pm$ 0.03 \\
& $M=500$  &-2.75 $\pm$ 0.01 & -2.26 $\pm$ 0.04 & -2.83 $\pm$ 0.00 &-5.38 $\pm$ 0.03  \\
\cmidrule(lr){1-6}
\multirow{4}{*}{\shortstack{Deep GP \\ $M=100$}} & $L=2$      & -2.75 $\pm$ 0.01  &-2.32 $\pm$ 0.03 & -2.81 $\pm$ 0.00 &-5.48 $\pm$ 0.04  \\
& $L=3$& -2.74 $\pm$ 0.01 & -2.28 $\pm$ 0.03 & -2.75 $\pm$ 0.00 &-5.42 $\pm$ 0.04  \\
& $L=4$      &-2.74 $\pm$ 0.01 & -2.25 $\pm$ 0.03 & -2.73 $\pm$ 0.00 &-5.36 $\pm$ 0.03   \\
& $L=5$      &-2.73 $\pm$ 0.01 &-2.24 $\pm$ 0.03  & -2.71 $\pm$ 0.00 &-5.34 $\pm$ 0.03   \\
\cmidrule(lr){1-6}
\multirow{5}{*}{\shortstack{DiffGP \\ $M=100$}}
&$T= 1.0$ & -2.75 $\pm$ 0.01 &-2.30 $\pm$ 0.03 & -2.79 $\pm$ 0.04 &-5.40 $\pm$ 0.04 \\ 
&$T= 2.0$ & -2.74 $\pm$ 0.01 & -2.27 $\pm$ 0.04 &-2.78 $\pm$ 0.04 &-5.36 $\pm$ 0.05\\ 
&$T= 3.0$ & -2.72 $\pm$ 0.01 &-2.25 $\pm$ 0.04 & -2.79 $\pm$ 0.00 &-5.31 $\pm$ 0.04\\ 
&$T= 4.0$ & -2.72 $\pm$ 0.01 &-2.25 $\pm$ 0.03 & -2.78 $\pm$ 0.00 &-5.29 $\pm$ 0.03\\ 
&$T= 5.0$ & -2.72 $\pm$ 0.01 &-2.26 $\pm$ 0.02 & -2.77 $\pm$ 0.00 &-5.28 $\pm$ 0.03\\ 
\cmidrule(lr){1-6}
\multirow{5}{*}{\shortstack{FB-DiffGP (ours) \\ $M=100$}}
&$T= 1.0$ & -\textbf{2.68} $\pm$ 0.01 &-\textbf{2.28} $\pm$ 0.04 & -\textbf{2.59} $\pm$ 0.02 &-\textbf{5.18} $\pm$ 0.04\\ 
&$T= 2.0$ & -\textbf{2.66} $\pm$ 0.01 &-\textbf{2.26} $\pm$ 0.04 & -\textbf{2.58} $\pm$ 0.02 &-\textbf{5.16} $\pm$ 0.03\\ 
&$T= 3.0$ & \textbf{-2.66} $\pm$ 0.01 &-\textbf{2.25} $\pm$ 0.03 & -\textbf{2.56} $\pm$ 0.01 &-\textbf{5.14} $\pm$ 0.03\\ 
&$T= 4.0$ &-\textbf{2.63} $\pm$ 0.02 & \textbf{-2.24} $\pm$ 0.04 & -\textbf{2.57} $\pm$ 0.01 &-\textbf{5.13} $\pm$ 0.04\\ 
&$T= 5.0$ & \textbf{-2.62} $\pm$ 0.04 & -\textbf{2.23} $\pm$ 0.03 &-\textbf{2.55} $\pm$ 0.01 &-\textbf{5.11} $\pm$ 0.03\\ 
\bottomrule
\end{tabular}
}
\caption{The test log-likelihood values on 8 benchmark datasets using 90\%/10\% random training and test splits with 20 repetitions.}
\label{LL}
\end{table}

\subsection{Regression benchmarks}
\label{rb}
We also compared our model with the state-of-the-art results from \cite{hegde2018deep} on eight regression benchmarks. Each experiment used 90\%/10\% random training and testing splits, with 20 repetitions. For both Gaussian Process methods, we used the RBF kernel with ARD and 100 inducing points. During testing, we computed the predictive mean and variance for each sample generated from Equation (\ref{forward}) and calculated the average summary statistics, including RMSE and Log Likelihood (LL), across these samples. The mean and standard error of RMSE are reported in Table \ref{rmse}, while the mean and standard error of LL are in Table \ref{LL}. From Table \ref{rmse} and Table \ref{LL}, it is clear that our method outperforms previous methods on all eight datasets, with significant improvements observed on the Boston, Energy, Concrete, and Protein datasets. The data were tested with flow time values ranging from 1 to 5. 
We also observed a similar trend as \cite{hegde2018deep}, where increasing the flow time can increase the model capacity without overfitting, consequently improving the model's generalization ability.

\begin{figure}[t!]
  \centering
  \includegraphics[width=0.6\textwidth]{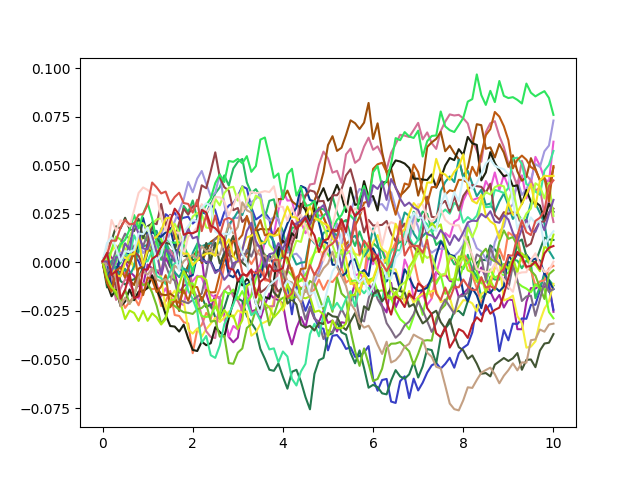}
  \caption{We display 30 sample paths to visualize the trajectories of the SDE for the Concrete data set. The plot shows that the posterior of our kernel hyperparameters is not a point estimate but a posterior SDE flow.  The diffusion process shown in this figure is  multivariate, but we only displayed a slice of it in the two-dimensional plot. The x-axis represents time $t$, while the y-axis represents the values of the hyperparameters.}
  \label{fig:example}
\end{figure}
\begin{figure}[t!]
   \centering
    \includegraphics[trim={8mm 0mm 10mm 0mm},clip,width=0.49\textwidth]{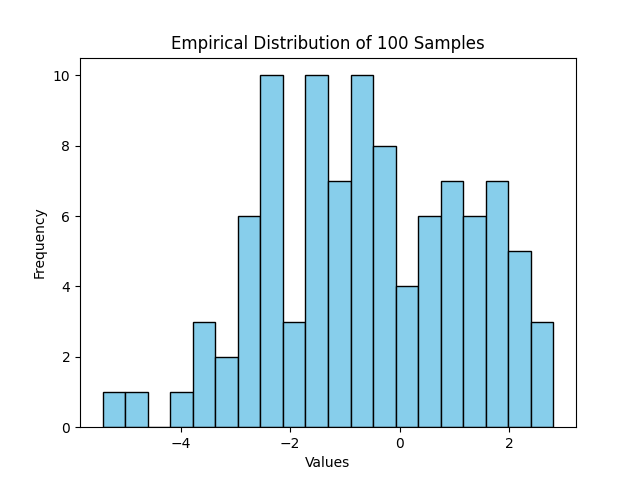}
    \includegraphics[trim={8mm 0mm 10mm 0mm},clip,width=0.49\textwidth]{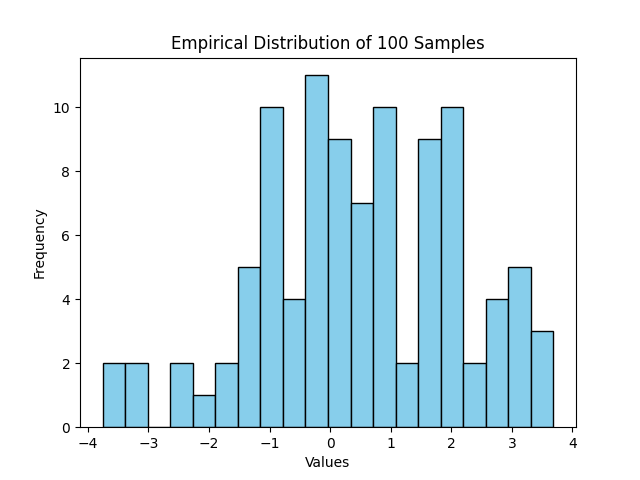}
    \caption{An empirical histogram of 100 sample paths at time $T=10.0$ for the Concrete (left) and Energy (right) datasets is shown. The x-axis represents the values of the kernel parameters, while the y-axis shows the empirical distribution. From the figure, it is clear that the learned posterior distribution of kernel parameters is no longer a point estimate, but instead takes the form of a probability distribution.}
    \label{fig2:example}
  \end{figure}

\paragraph{Simulation of kernel hyperparameter flows} To illustrate the improvement over the baseline more clearly, we simulated the trajectory of the learned kernel hyperparameters $\boldsymbol{\lambda}$ SDE. Since $\boldsymbol{\lambda}$ is indeed multivariate, we only displayed a slice of it. For visualization purposes, we selected the Concrete dataset and set $T = 10.0$ . The results are shown in Figure \ref{fig:example}. The x-axis represents time $t$, while the y-axis represents the values of the hyperparameters. In order to facilitate the visualization of the trajectories of the SDE, we displayed 30 sample paths in the figure and selected fixed initial values for optimization purposes. From the plot, we can observe that the posterior of our kernel hyperparameters is no longer a point estimate but a posterior SDE flow. Our trajectories serve to illustrate what the posterior SDE has learned, providing insight into the uncertainty estimation for the hyperparameters. The detailed uncertainty estimates are provided in Table \ref{LL}. In Figure \ref{fig2:example}, we display the empirical distribution of these 100 paths at $T=10$ for two data sets.


\begin{table*}[t!]

\resizebox{1.00\textwidth}{!}{
\begin{tabular}{ l r  cc cc cc cc }
\toprule
& & Yacht & Boston & Energy & Concrete    & Power  & Elevators & Protein & Year \\
\cmidrule(lr){3-10}
& $N$ &308& 506 & 768 & 1,030   & 9K   & 16K & 45K & 515K  \\
& $D$ &6& 13 & 8 & 8   & 4 & 18 & 9 &90 \\
\cmidrule(lr){1-10}
\shortstack{DiffGP}
& &0.59&0.61& 0.62& 0.70& 2.34& 3.85& 11.7&120.6\\ 
\cmidrule(lr){1-10}
\multirow{1}{*}{\shortstack{FB-DiffGP}}
&  &0.61&0.64 & 0.66 & 0.75 & 2.45   &  4.03 &12.9 &126.9\\ 

\bottomrule
\end{tabular}
}
\caption{Runtime (in seconds) of the proposed algorithm compared to the baseline for $M=100$ and $T=1$. Shown is the time required to complete one full pass of the entire dataset, also known as one epoch. The results show that FB-DiffGP does not lead to a substantial increase in computational costs, owing to the efficient parallel processing capabilities of deep learning GPUs.}
\label{time}
\end{table*}

\paragraph{Computational efficiency} Similar to the baseline algorithms, each training iteration of FB-DiffGP involves computing the inverse covariance with a complexity of $\mathcal{O}(M^3)$. In Table \ref{time}, we compare our method with the baseline DiffGP trained for a fixed number of epochs. The experiment is repeated five times on the same fold, and the results are averaged. Each run is conducted on a dedicated instance in a cloud computing platform equipped with a single Tesla A100 GPU and an Intel Core i9-13900K CPU. The results demonstrate that FB-DiffGP does not significantly increase computational costs, thanks to the efficient parallel processing capabilities of deep learning GPUs.

\paragraph{Comparison with different numbers of inducing points} We compared the performance of our algorithm with \(M=100\) and \(M=500\) inducing points, shown in Table \ref{M}. Our findings indicate that, similar to classical sparse GP methods, the performance improves with a larger number of inducing points. However, this improvement comes at the cost of increased training time, as the computational complexity grows with the number of inducing points.

\begin{table*}[t!]
\resizebox{1.00\textwidth}{!}{
\begin{tabular}{ l r  cc cc }
\toprule
& & Yacht & Boston & Energy & Concrete    \\
\cmidrule(lr){3-6}
& $N\,|\,D$ &308 $|$ 6& 506 $|$ 13 & 768 $|$ 8 & 1,030 $|$ 8    \\
\cmidrule(lr){1-6}
\shortstack{FB-DiffGP $M=100$}
& &0.43 $\pm$ 0.04&2.63 $\pm$ 0.6 & 0.42 $\pm$ 0.01 &4.75 $\pm$ 0.12 \\ 
\cmidrule(lr){1-6}
\multirow{1}{*}{\shortstack{FB-DiffGP $M=500$}}
&  &0.42 $\pm$ 0.04 &2.52 $\pm$ 0.07 & 0.40 $\pm$ 0.01 &4.46 $\pm$ 0.14 \\ 
\bottomrule
\toprule
& &  Power  & Elevators & Protein & Year \\
\cmidrule(lr){3-6}
& $N\,|\,D$ &9,568 $|$ 4   & 16,599 $|$ 18 & 45,730 $|$ 9 &515,345 $|$ 90  \\
\cmidrule(lr){1-6}
\shortstack{FB-DiffGP $M=100$}
& &3.64 $\pm$ 0.03 &4.32 $\pm$ 0.03 & 3.94 $\pm$ 0.03&5.40 $\pm$ 0.04\\ 
\cmidrule(lr){1-6}
\multirow{1}{*}{\shortstack{FB-DiffGP $M=500$}}
&  &3.62 $\pm$ 0.03 &4.29 $\pm$ 0.03 & 3.86 $\pm$ 0.03&5.38 $\pm$ 0.03\\ 
\bottomrule
\end{tabular}
}
\caption{The RMSE of the proposed algorithm with varying numbers of inducing points \(M\) is shown. As the number of inducing points increases, the performance improves, leading to lower RMSE values.}
\label{M}
\end{table*} 



\paragraph{Comparison with Recent Deep Gaussian Process Baselines}
To further demonstrate our proposed method, we compare with the IPVI \cite{yu2019implicit} and NOVI \cite{xu2023neural} models, which are recent deep Gaussian process models that also utilize neural networks for inference. We show comparisons using the UCI datasets, evaluating the models based on RMSE (Table \ref{rmse1}). The results show that our FB-DiffGP method remains competitive when compared to these recent state-of-the-art methods.

\begin{table}[thp!]
\centering
\resizebox{.9\textwidth}{!}{
\begin{tabular}{ l r  c c c c }
\toprule
& & Yacht & Boston & Energy & Concrete     \\
\cmidrule(lr){3-6}
& $N\,|\,D$ &308 $|$ 6& 506 $|$ 13 & 768 $|$ 8 & 1,030 $|$ 8  \\

\cmidrule(lr){1-6}
\multirow{4}{*}{\shortstack{IPVI \\ $M=100$}} & $L=2$ & 0.43 $\pm$ 0.04 & 2.95 $\pm$ 0.15  & 0.67 $\pm$ 0.03 &5.32 $\pm$ 0.15 \\
& $L=3$&  0.43 $\pm$ 0.04 &2.94 $\pm$ 0.14  & 0.65 $\pm$ 0.04 &5.28 $\pm$ 0.14  \\
& $L=4$      &   0.44 $\pm$ 0.03 & 2.92 $\pm$ 0.13 & 0.64 $\pm$ 0.02 &5.29 $\pm$ 0.12  \\
& $L=5$      &  0.42 $\pm$ 0.02 &2.90 $\pm$ 0.13  & 0.63 $\pm$ 0.02 &5.27 $\pm$ 0.11 \\
\cmidrule(lr){1-6}
\multirow{4}{*}{\shortstack{NOVI \\ $M=100$}} & $L=2$ & 0.43 $\pm$ 0.03 & 2.65 $\pm$ 0.14  & 0.48 $\pm$ 0.03 &4.83 $\pm$ 0.13 \\
& $L=3$&  0.46 $\pm$ 0.03 &2.70 $\pm$ 0.14  & 0.46 $\pm$ 0.04 &4.85 $\pm$ 0.12  \\
& $L=4$      &   0.45 $\pm$ 0.04 & 2.76 $\pm$ 0.13 & 0.44 $\pm$ 0.02 &4.80 $\pm$ 0.12  \\
& $L=5$      &  0.43 $\pm$ 0.03 &2.74 $\pm$ 0.11  & 0.42 $\pm$ 0.01 &4.78 $\pm$ 0.09 \\

\cmidrule(lr){1-6}
\multirow{5}{*}{\shortstack{FB-DiffGP (ours) \\ $M=100$}}
& $T= 1.0$ &0.43 $\pm$ 0.04 &2.63 $\pm$ 0.10 & 0.42 $\pm$ 0.01 & 4.75 $\pm$ 0.12  \\ 
& $T= 2.0$ &0.41 $\pm$ 0.03 & 2.49 $\pm$ 0.10 & 0.41$\pm$ 0.02 & 4.33 $\pm$ 0.11 \\
& $T= 3.0$ &0.41 $\pm$ 0.03 & 2.47 $\pm$ 0.11 & 0.39 $\pm$ 0.01 &4.22 $\pm$ 0.12  \\
& $T= 4.0$ &0.40 $\pm$ 0.02 & 2.45 $\pm$ 0.09 & 0.37 $\pm$ 0.02 & 4.07 $\pm$ 0.11  \\
& $T= 5.0$ &\textbf{0.38} $\pm$ 0.04 & \textbf{2.39} $\pm$ 0.10 & \textbf{0.38} $\pm$ 0.01 & \textbf{4.01} $\pm$ 0.11  \\
\bottomrule
\toprule
& &  Power  & Elevators & Protein & Year  \\
\cmidrule(lr){3-6}
& $N\,|\,D$ & 9,568 $|$ 4   & 16,599 $|$ 18 & 45,730 $|$ 9 &515,345 $|$ 90  \\

\cmidrule(lr){1-6}
\multirow{4}{*}{\shortstack{IPVI \\ $M=100$}} & $L=2$ & 3.78 $\pm$ 0.03 & 4.38 $\pm$ 0.04  & 3.98 $\pm$ 0.03 &5.44 $\pm$ 0.04 \\
& $L=3$&  3.74 $\pm$ 0.03 &4.35 $\pm$ 0.03  & 3.77 $\pm$ 0.04 &5.38 $\pm$ 0.04  \\
& $L=4$      &   3.70 $\pm$ 0.03 & 4.32 $\pm$ 0.03 & 3.75 $\pm$ 0.04 &5.34 $\pm$ 0.03  \\
& $L=5$      &  3.68 $\pm$ 0.03 & 4.30 $\pm$ 0.03  &3.72 $\pm$ 0.03 &5.31$\pm$ 0.02 \\
\cmidrule(lr){1-6}
\multirow{4}{*}{\shortstack{NOVI \\ $M=100$}} & $L=2$ & 3.79 $\pm$ 0.03 & 4.34 $\pm$ 0.04  & 3.95 $\pm$ 0.03 &5.42 $\pm$ 0.04 \\
& $L=3$&  3.75 $\pm$ 0.04 &4.33 $\pm$ 0.04  & 3.81 $\pm$ 0.03 &5.38 $\pm$ 0.03  \\
& $L=4$      &   3.73 $\pm$ 0.03 & 4.31 $\pm$ 0.03 & 3.74 $\pm$ 0.03 &5.32 $\pm$ 0.03  \\
& $L=5$      &  3.70 $\pm$ 0.03 &4.29 $\pm$ 0.02  & \textbf{3.71} $\pm$ 0.03 &5.28 $\pm$ 0.03 \\

\cmidrule(lr){1-6}
\multirow{5}{*}{\shortstack{FB-DiffGP (ours) \\ $M=100$}}
& $T= 1.0$ &    3.64 $\pm$ 0.03 &4.32 $\pm$ 0.03 &3.94 $\pm$ 0.03 &5.40 $\pm$ 0.04 \\ 
& $T= 2.0$ &     3.61 $\pm$ 0.03 & 4.30 $\pm$ 0.03 &3.88 $\pm$ 0.03 &5.36 $\pm$ 0.03\\
& $T= 3.0$ &    3.58 $\pm$ 0.03 & 4.28 $\pm$ 0.04 & 3.85 $\pm$ 0.03 &5.33 $\pm$ 0.03  \\
& $T= 4.0$ & 3.54 $\pm$ 0.03 & 4.25 $\pm$ 0.03 & 3.81 $\pm$ 0.03 &5.28 $\pm$ 0.03 \\
& $T= 5.0$ &  \textbf{3.53} $\pm$ 0.03 &\textbf{4.25} $\pm$ 0.02 & 3.79 $\pm$ 0.03 &\textbf{5.24} $\pm$ 0.03  \\
\bottomrule
\end{tabular}
}
\caption{The test RMSE values on 8 benchmark datasets, using 90\%/10\% random training and test splits with 20 repetitions, compared with recent deep Gaussian process baselines. }
\label{rmse1}
\end{table}
\subsection{Classification Benchmarks}
\label{cb}

We performed large-scale experiments using the Higgs dataset, which contains 11 million data points with 28 features. This dataset was created through Monte Carlo simulations modeling particle dynamics in accelerators for Higgs boson detection. We randomly split the data, using 90\% for training and the remaining 10\% for testing. To evaluate the performance, we use the area under the curve (AUC) metric and compared our results with previously reported methods. Table \ref{class} presents the obtained test performance, demonstrating that our FB-DiffGP method outperforms the competing approaches. Additionally, we conducted experiments on the SUSY dataset, and the results showcased the competitive performance of our proposed algorithm.

\begin{table}[th]
\centering
\resizebox{.85\width}{!}{
\begin{tabular}{ l r c c }
\toprule
 & & SUSY & HIGGS \\
\cmidrule(lr){3-4}
 & $N\,|\,D$ & 5,500,000 $\,|\,$ 18 & 11,000,000 $\,|\,$ 28\\
 \midrule
\midrule
DNN & & 0.876 & \textbf{0.885}\\
\multirow{2}{*}{Sparse GP} & $M=100$ & 0.875 & 0.785 \\
& $M=500$ & 0.876 & 0.794 \\
\midrule
\multirow{4}{*}{\shortstack{Deep GP \\ $M=100$}} & $L=2$      & 0.877 & 0.830 \\
& $L=3$      & 0.877 & 0.837 \\
& $L=4$      & 0.877 & 0.841 \\
& $L=5$      & 0.877 & 0.846 \\
\midrule
\multirow{2}{*}{\shortstack{DiffGP \\ $M=100$}} 
& $T=1.0$ & 0.878 & 0.840 \\
& $T=3.0$ & 0.878& 0.841 \\
& $T=5.0$ & 0.878 & 0.842 \\
\midrule
\multirow{2}{*}{\shortstack{FB-DiffGP(ours) \\ $M=100$}} 
& $T=1.0$ & \textbf{0.887} & 0.852 \\
& $T=3.0$ & \textbf{0.887} & 0.856 \\
& $T=5.0$ & \textbf{0.887} & 0.857 \\
\bottomrule
\end{tabular}
}
\caption{The test AUC values for large-scale classification datasets, using a 90\% / 10\% random split for training and testing, show that our method overall outperforms the baseline, further demonstrating the scalability of the approach.}
\label{class}
\end{table}

\section{Conclusion and Future Work}
\label{section6}
We have proposed a fully Bayesian approach to Gaussian Process (GP) modeling, where kernel hyperparameters are treated as random variables, and interconnected stochastic differential equations  are used to infer the posterior distribution of DiffGPs. By capturing uncertainty in hyperparameter estimation, our method significantly enhances the model's adaptability to complex system dynamics. Experimental results demonstrate that our approach outperforms conventional techniques, showing improved accuracy.

However, there are limitations to our method, including the optimization of hyperparameters such as the SDE time \(T\), the number of inducing points \(M\), and the architecture of the neural SDE. These require additional methods, such as cross-validation or Bayesian optimization, to tune effectively. In our current experiments, we primarily relied on empirical methods. Furthermore, balancing the trade-off between accuracy and computational efficiency remains an open challenge, which we plan to address in future work. This research also can generalize to other problems involving continuous-time Gaussian processes, e.g., \cite{paisley2022bayesian}. Future exploration includes expanding the application of our FB-DiffGP model in diverse domains such as spatio-temporal model averaging, image analysis, and financial datasets.

\bibliographystyle{elsarticle-num} 
\bibliography{ref}

@article{hegde2018deep,
  title={Deep learning with differential {G}aussian process flows},
  author={Hegde, Pashupati and Heinonen, Markus and L{\"a}hdesm{\"a}ki, Harri and Kaski, Samuel},
  journal={arXiv preprint arXiv:1810.04066},
  year={2018}
}

@article{rouard2021crash,
  title={CRASH: Raw audio score-based generative modeling for controllable high-resolution drum sound synthesis},
  author={Rouard, Simon and Hadjeres, Ga{\"e}tan},
  journal={arXiv preprint arXiv:2106.07431},
  year={2021}
}

@inproceedings{tran2021sparse,
  title={Sparse within sparse gaussian processes using neighbor information},
  author={Tran, Gia-Lac and Milios, Dimitrios and Michiardi, Pietro and Filippone, Maurizio},
  booktitle={International Conference on Machine Learning},
  pages={10369--10378},
  year={2021},
  organization={PMLR}
}

@article{jafrasteh2021input,
  title={Input dependent sparse gaussian processes},
  author={Jafrasteh, Bahram and Villacampa-Calvo, Carlos and Hern{\'a}ndez-Lobato, Daniel},
  journal={arXiv preprint arXiv:2107.07281},
  year={2021}
}

@article{song2020score,
  title={Score-based generative modeling through stochastic differential equations},
  author={Song, Yang and Sohl-Dickstein, Jascha and Kingma, Diederik P and Kumar, Abhishek and Ermon, Stefano and Poole, Ben},
  journal={arXiv preprint arXiv:2011.13456},
  year={2020}
}

@article{hong2023debiasing,
  title={Debiasing scores and prompts of 2d diffusion for view-consistent text-to-3d generation},
  author={Hong, Susung and Ahn, Donghoon and Kim, Seungryong},
  journal={Advances in Neural Information Processing Systems},
  volume={36},
  pages={11970--11987},
  year={2023}
}

@article{ma2022accelerating,
  title={Accelerating score-based generative models for high-resolution image synthesis},
  author={Ma, Hengyuan and Zhang, Li and Zhu, Xiatian and Zhang, Jingfeng and Feng, Jianfeng},
  journal={arXiv preprint arXiv:2206.04029},
  year={2022}
}

@misc{higgs_280,
  author       = {Whiteson, Daniel},
  title        = {{HIGGS}},
  year         = {2014},
  howpublished = {UCI Machine Learning Repository},
  note         = {{DOI}: https://doi.org/10.24432/C5V312}
}

@misc{susy_279,
  author       = {Whiteson, Daniel},
  title        = {{SUSY}},
  year         = {2014},
  howpublished = {UCI Machine Learning Repository},
  note         = {{DOI}: https://doi.org/10.24432/C54606}
}

@misc{asuncion2007uci,
  title={UCI machine learning repository},
  author={Asuncion, Arthur and Newman, David and others},
  year={2007},
  publisher={Irvine, CA, USA}
}

@article{bishop1993analysis,
  title={Analysis of multiphase flows using dual-energy gamma densitometry and neural networks},
  author={Bishop, Christopher M and James, Gwilym D},
  journal={Nuclear Instruments and Methods in Physics Research Section A: Accelerators, Spectrometers, Detectors and Associated Equipment},
  volume={327},
  number={2-3},
  pages={580--593},
  year={1993},
  publisher={Elsevier}
}

@article{paisley2010active,
  title={Active learning and basis selection for kernel-based linear models: A {B}ayesian perspective},
  author={Paisley, John and Liao, Xuejun and Carin, Lawrence},
  journal={IEEE Transactions on Signal Processing},
  volume={58},
  number={5},
  pages={2686--2700},
  year={2010},
  publisher={IEEE}
}

@inproceedings{sun2017location,
  title={Location dependent {D}irichlet processes},
  author={Sun, Shiliang and Paisley, John and Liu, Qiuyang},
  booktitle={Conference on Intelligence Science and Big Data Engineering},
  year={2017}
}

@inproceedings{liang2015landmarking,
  title={Landmarking manifolds with {G}aussian processes},
  author={Liang, Dawen and Paisley, John},
  booktitle={International Conference on Machine Learning},
  year={2015}
}

@inproceedings{titsias2010bayesian,
  title={Bayesian {G}aussian process latent variable model},
  author={Titsias, Michalis and Lawrence, Neil D},
  booktitle={Artificial Intelligence and Statistics},
  year={2010}}

@inproceedings{xu2024sparse,
  title={Sparse Inducing Points in Deep {G}aussian Processes: {E}nhancing Modeling with Denoising Diffusion Variational Inference},
  author={Xu, Jian and Zeng, Delu and Paisley, John},
  booktitle={International Conference on Machine Learning},
  year={2024}
}

@article{baldi2014searching,
  title={Searching for exotic particles in high-energy physics with deep learning},
  author={Baldi, Pierre and Sadowski, Peter and Whiteson, Daniel},
  journal={Nature Communications},
  volume={5},
  number={1},
  pages={4308},
  year={2014},
  publisher={Nature Publishing Group UK London}
}

@incollection{rasmussen2003gaussian,
  title={Gaussian processes in machine learning},
  author={Rasmussen, Carl Edward},
  booktitle={Summer School on Machine Learning},
  pages={63--71},
  year={2003},
  publisher={Springer}
}

@inproceedings{hensman2015mcmc,
  title={{MCMC} for variationally sparse {G}aussian processes},
  author={Hensman, James and Matthews, Alexander G and Filippone, Maurizio and Ghahramani, Zoubin},
  booktitle={Advances in Neural Information Processing Systems},
  year={2015}
}

@inproceedings{hensman2015scalable,
  title={Scalable variational {G}aussian process classification},
  author={Hensman, James and Matthews, Alexander and Ghahramani, Zoubin},
  booktitle={Artificial Intelligence and Statistics},
  year={2015},
}

@inproceedings{damianou2013deep,
  title={Deep {G}aussian processes},
  author={Damianou, Andreas and Lawrence, Neil D},
  booktitle={Artificial Intelligence and Statistics},
  year={2013}}

@article{salimbeni2017doubly,
  title={Doubly stochastic variational inference for deep {G}aussian processes},
  author={Salimbeni, Hugh and Deisenroth, Marc},
  journal={Advances in Neural Information Processing Systems},
  year={2017}
}

@inproceedings{yildiz2018learning,
  title={Learning stochastic differential equations with {G}aussian processes without gradient matching},
  author={Yildiz, Cagatay and Heinonen, Markus and Intosalmi, Jukka and Mannerstrom, Henrik and Lahdesmaki, Harri},
  booktitle={Workshop on Machine Learning for Signal Processing},
  year={2018}
}

@inproceedings{lalchand2020approximate,
  title={Approximate inference for fully {B}ayesian {G}aussian process regression},
  author={Lalchand, Vidhi and Rasmussen, Carl Edward},
  booktitle={Symposium on Advances in Approximate Bayesian Inference},
  year={2020},
}

@inproceedings{lalchand2022sparse,
  title={Sparse {G}aussian process hyperparameters: Optimize or Integrate?},
  author={Lalchand, Vidhi and Bruinsma, Wessel and Burt, David and Rasmussen, Carl Edward},
  booktitle={Advances in Neural Information Processing Systems},
  year={2022}
}

@inproceedings{zhang2019anodev2,
  title={{ANODEV}2: A coupled neural {ODE} framework},
  author={Zhang, Tianjun and Yao, Zhewei and Gholami, Amir and Gonzalez, Joseph E and Keutzer, Kurt and Mahoney, Michael W and Biros, George},
  booktitle={Advances in Neural Information Processing Systems},
  year={2019}
}

@inproceedings{xu2022infinitely,
  title={Infinitely deep {B}ayesian neural networks with stochastic differential equations},
  author={Xu, Winnie and Chen, Ricky TQ and Li, Xuechen and Duvenaud, David},
  booktitle={Artificial Intelligence and Statistics},
  year={2022}
}

@article{stanley2009hypercube,
  title={A hypercube-based encoding for evolving large-scale neural networks},
  author={Stanley, Kenneth O and D'Ambrosio, David B and Gauci, Jason},
  journal={Artificial Life},
  volume={15},
  number={2},
  pages={185--212},
  year={2009}
}

@inproceedings{koutnik2010evolving,
  title={Evolving neural networks in compressed weight space},
  author={Koutnik, Jan and Gomez, Faustino and Schmidhuber, J{\"u}rgen},
  booktitle={Conference on Genetic and Evolutionary Computation},
  year={2010}
}

@article{DBLP:journals/corr/HaDL16,
  author       = {David Ha and
                  Andrew M. Dai and
                  Quoc V. Le},
  title        = {HyperNetworks},
  journal      = {CoRR},
  volume       = {abs/1609.09106},
  year         = {2016},
  url          = {http://arxiv.org/abs/1609.09106},
  eprinttype    = {arXiv},
  eprint       = {1609.09106},
  bibsource    = {dblp computer science bibliography, https://dblp.org}
}

@inproceedings{li2020scalable,
  title={Scalable gradients for stochastic differential equations},
  author={Li, Xuechen and Wong, Ting-Kam Leonard and Chen, Ricky TQ and Duvenaud, David},
  booktitle={Artificial Intelligence and Statistics},
  year={2020}
}

@article{opper2019variational,
  title={Variational inference for stochastic differential equations},
  author={Opper, Manfred},
  journal={Annalen der Physik},
  volume={531},
  number={3},
  year={2019}
}

@inproceedings{agrawal2021amortized,
  title={Amortized variational inference for simple hierarchical models},
  author={Agrawal, Abhinav and Domke, Justin},
  booktitle={Advances in Neural Information Processing Systems},
  year={2021}
}

@inproceedings{kim2018semi,
  title={Semi-amortized variational autoencoders},
  author={Kim, Yoon and Wiseman, Sam and Miller, Andrew and Sontag, David and Rush, Alexander},
  booktitle={International Conference on Machine Learning},
  year={2018}
}

@inproceedings{kidger2021neural,
  title={Neural {SDE}s as infinite-dimensional {GAN}s},
  author={Kidger, Patrick and Foster, James and Li, Xuechen and Lyons, Terry J},
  booktitle={International conference on Machine Learning},
  year={2021}
}

@article{boue1998variational,
  title={A variational representation for certain functionals of {B}rownian motion},
  author={Bou{\'e}, Michelle and Dupuis, Paul},
  journal={The Annals of Probability},
  volume={26},
  number={4},
  pages={1641--1659},
  year={1998}
}

@article{tzen2019neural,
  title={Neural stochastic differential equations: Deep latent gaussian models in the diffusion limit},
  author={Tzen, Belinda and Raginsky, Maxim},
  journal={arXiv preprint arXiv:1905.09883},
  year={2019}
}

@article{amari1993backpropagation,
  title={Backpropagation and stochastic gradient descent method},
  author={Amari, Shun-ichi},
  journal={Neurocomputing},
  volume={5},
  number={4-5},
  pages={185--196},
  year={1993}
}

@article{hoffman2013stochastic,
  title={Stochastic variational inference},
  author={Hoffman, Matthew D and Blei, David M and Wang, Chong and Paisley, John},
  journal={Journal of Machine Learning Research},
number = {14},
 pages = {1303-1347},
  year={2013}
}

@article{matthews2017gpflow,
  title={{GP}flow: A {G}aussian Process Library using {T}ensorFlow.},
  author={Matthews, Alexander G de G and Van Der Wilk, Mark and Nickson, Tom and Fujii, Keisuke and Boukouvalas, Alexis and Le{\'o}n-Villagr{\'a}, Pablo and Ghahramani, Zoubin and Hensman, James},
  journal={Journal of Machine Learning Research},
  volume={18},
  number={40},
  pages={1--6},
  year={2017}
}

@article{krauth2016autogp,
  title={AutoGP: Exploring the capabilities and limitations of Gaussian process models},
  author={Krauth, Karl and Bonilla, Edwin V and Cutajar, Kurt and Filippone, Maurizio},
  journal={arXiv preprint arXiv:1610.05392},
  year={2016}
}

@inproceedings{gardner2018gpytorch,
  title={Gpytorch: Blackbox matrix-matrix gaussian process inference with gpu acceleration},
  author={Gardner, Jacob and Pleiss, Geoff and Weinberger, Kilian Q and Bindel, David and Wilson, Andrew G},
  booktitle={Advances in Neural Information Processing Systems},
  year={2018}
}

@inproceedings{titsias2009variational,
  title={Variational learning of inducing variables in sparse {G}aussian processes},
  author={Titsias, Michalis},
  booktitle={Artificial Intelligence and Statistics},
  year={2009}
}

@inproceedings{lazaro2009inter,
  title={Inter-domain {G}aussian processes for sparse inference using inducing features},
  author={L{\'a}zaro-Gredilla, Miguel and Figueiras-Vidal, Anibal},
  booktitle={Advances in Neural Information Processing Systems},
  year={2009}
}

@inproceedings{van2017convolutional,
  title={Convolutional {G}aussian processes},
  author={Van der Wilk, Mark and Rasmussen, Carl Edward and Hensman, James},
  booktitle={Advances in Neural Information Processing Systems},
  year={2017}
}

@article{li2016review,
  title={A review on {G}aussian process latent variable models},
  author={Li, Ping and Chen, Songcan},
  journal={CAAI Transactions on Intelligence Technology},
  volume={1},
  number={4},
  pages={366--376},
  year={2016}
}

@inproceedings{corani2021time,
  title={Time series forecasting with {G}aussian processes needs priors},
  author={Corani, Giorgio and Benavoli, Alessio and Zaffalon, Marco},
  booktitle={Joint European Conference on Machine Learning and Knowledge Discovery in Databases},
  year={2021}
}

@inproceedings{schreiter2015sparse,
  title={Sparse {G}aussian process regression for compliant, real-time robot control},
  author={Schreiter, Jens and Englert, Peter and Nguyen-Tuong, Duy and Toussaint, Marc},
  booktitle={IEEE International Conference on Robotics and Automation},
  year={2015}
}

@inproceedings{tran2019calibrating,
  title={Calibrating deep convolutional {G}aussian processes},
  author={Tran, Gia-Lac and Bonilla, Edwin V and Cunningham, John and Michiardi, Pietro and Filippone, Maurizio},
  booktitle={Artificial Intelligence and Statistics},
  year={2019}
}

@article{bernardo1998regression,
  title={Regression and classification using {G}aussian process priors},
  author={Bernardo, J and Berger, J and Dawid, APAFMS and Smith, A and others},
  journal={Bayesian Statistics},
  volume={6},
  pages={475},
  year={1998}
}

@inproceedings{williams1995gaussian,
  title={Gaussian processes for regression},
  author={Williams, Christopher and Rasmussen, Carl},
  booktitle={Advances in Neural Information Processing Systems},
  year={1995}
}

@inproceedings{barber1996gaussian,
  title={Gaussian processes for {B}ayesian classification via hybrid {M}onte {C}arlo},
  author={Barber, David and Williams, Christopher},
  booktitle={Advances in Neural Information Processing Systems},
  year={1996}
}

@inproceedings{murray2010slice,
  title={Slice sampling covariance hyperparameters of latent {G}aussian models},
  author={Murray, Iain and Adams, Ryan P},
  booktitle={Advances in Neural Information Processing Systems},
  year={2010}
}

@article{bui2018partitioned,
  title={Partitioned variational inference: {A} unified framework encompassing federated and continual learning},
  author={Bui, Thang D and Nguyen, Cuong V and Swaroop, Siddharth and Turner, Richard E},
  journal={arXiv preprint arXiv:1811.11206},
  year={2018}
}

@inproceedings{rossi2021sparse,
  title={Sparse {G}aussian processes revisited: {B}ayesian approaches to inducing-variable approximations},
  author={Rossi, Simone and Heinonen, Markus and Bonilla, Edwin and Shen, Zheyang and Filippone, Maurizio},
  booktitle={Artificial Intelligence and Statistics},
  year={2021}
}

@inproceedings{bui2017streaming,
  title={Streaming sparse {G}aussian process approximations},
  author={Bui, Thang D and Nguyen, Cuong and Turner, Richard E},
  booktitle={Advances in Neural Information Processing Systems},
  year={2017}
}

@article{liu2019neural,
  title={Neural {SDE}: Stabilizing neural ode networks with stochastic noise},
  author={Liu, Xuanqing and Xiao, Tesi and Si, Si and Cao, Qin and Kumar, Sanjiv and Hsieh, Cho-Jui},
  journal={arXiv preprint arXiv:1906.02355},
  year={2019}
}

@inproceedings{liu2020does,
  title={How does noise help robustness? {E}xplanation and exploration under the neural {SDE} framework},
  author={Liu, Xuanqing and Xiao, Tesi and Si, Si and Cao, Qin and Kumar, Sanjiv and Hsieh, Cho-Jui},
  booktitle={Computer Vision and Pattern Recognition},
  year={2020}
}

@article{kong2020sde,
  title={{SDE}-{NET}: {E}quipping deep neural networks with uncertainty estimates},
  author={Kong, Lingkai and Sun, Jimeng and Zhang, Chao},
  journal={arXiv preprint arXiv:2008.10546},
  year={2020}
}

@inproceedings{kidger2021efficient,
  title={Efficient and accurate gradients for neural {SDE}s},
  author={Kidger, Patrick and Foster, James and Li, Xuechen Chen and Lyons, Terry},
  booktitle={Advances in Neural Information Processing Systems},
  year={2021}
}

@article{yang2023neural,
  title={Neural network stochastic differential equation models with applications to financial data forecasting},
  author={Yang, Luxuan and Gao, Ting and Lu, Yubin and Duan, Jinqiao and Liu, Tao},
  journal={Applied Mathematical Modeling},
  volume={115},
  pages={279--299},
  year={2023},
}

@inproceedings{ho2020denoising,
  title={Denoising diffusion probabilistic models},
  author={Ho, Jonathan and Jain, Ajay and Abbeel, Pieter},
  booktitle={Advances in Neural Information Processing Systems},
  year={2020}
}

@article{chen2019incremental,
  title={Incremental Reinforcement Learning---a New Continuous Reinforcement Learning Frame Based on Stochastic Differential Equation methods},
  author={Chen, Tianhao and Cheng, Limei and Liu, Yang and Jia, Wenchuan and Ma, Shugen},
  journal={arXiv preprint arXiv:1908.02974},
  year={2019}
}

@inproceedings{havasi2018inference,
  title={Inference in deep {G}aussian processes using stochastic gradient Hamiltonian Monte Carlo},
  author={Havasi, Marton and Hern{\'a}ndez-Lobato, Jos{\'e} Miguel and Murillo-Fuentes, Juan Jos{\'e}},
  booktitle={Advances in Neural Information Processing Systems},
  year={2018}
}

@inproceedings{yu2019implicit,
  title={Implicit posterior variational inference for deep {G}aussian processes},
  author={Yu, Haibin and Chen, Yizhou and Low, Bryan Kian Hsiang and Jaillet, Patrick and Dai, Zhongxiang},
  booktitle={Advances in Neural Information Processing Systems},
  year={2019}
}

@article{xu2023neural,
  title={Neural Operator Variational Inference based on Regularized {S}tein Discrepancy for Deep {G}aussian Processes},
  author={Xu, Jian and Du, Shian and Yang, Junmei and Ma, Qianli and Zeng, Delu},
  journal={arXiv preprint arXiv:2309.12658},
  year={2023}
}

@inproceedings{solin2021scalable,
  title={Scalable inference in {SDE}s by direct matching of the Fokker--Planck--Kolmogorov equation},
  author={Solin, Arno and Tamir, Ella and Verma, Prakhar},
  booktitle={Advances in Neural Information Processing Systems},
  year={2021}
}

@inproceedings{rezende2014stochastic,
  title={Stochastic backpropagation and approximate inference in deep generative models},
  author={Rezende, Danilo Jimenez and Mohamed, Shakir and Wierstra, Daan},
  booktitle={International Conference on Machine Learning},
  year={2014}
}

@article{van1976stochastic,
  title={Stochastic differential equations},
  author={Van Kampen, Nicolaas G},
  journal={Physics Reports},
  volume={24},
  number={3},
  pages={171--228},
  year={1976}
}

@article{dunlop2018deep,
  title={How deep are deep {G}aussian processes?},
  author={Dunlop, Matthew M and Girolami, Mark A and Stuart, Andrew M and Teckentrup, Aretha L},
  journal={Journal of Machine Learning Research},
  volume={19},
  number={54},
  pages={1--46},
  year={2018}
}

@inproceedings{duvenaud2014avoiding,
  title={Avoiding pathologies in very deep networks},
  author={Duvenaud, David and Rippel, Oren and Adams, Ryan and Ghahramani, Zoubin},
  booktitle={Artificial Intelligence and Statistics},
  year={2014}
}

@inproceedings{paisley2022bayesian,
  title={Bayesian nonparametric model averaging using scalable {G}aussian process representations},
  author={Paisley, John and Rowland, Sebastian and Liu, Jeremiah Zhe and Coull, Brent and Kioumourtzoglou, Marianthi-Anna},
  booktitle={IEEE International Conference on Big Data},
  year={2022}
}

@article{jin2024forecasting,
  title={Forecasting wholesale prices of yellow corn through the Gaussian process regression},
  author={Jin, Bingzi and Xu, Xiaojie},
  journal={Neural Computing and Applications},
  volume={36},
  number={15},
  pages={8693--8710},
  year={2024},
  publisher={Springer}
}

@inproceedings{he2016deep,
  title={Deep residual learning for image recognition},
  author={He, Kaiming and Zhang, Xiangyu and Ren, Shaoqing and Sun, Jian},
  booktitle={Proceedings of the IEEE conference on computer vision and pattern recognition},
  pages={770--778},
  year={2016}
}

@article{chen2018neural,
  title={Neural ordinary differential equations},
  author={Chen, Ricky TQ and Rubanova, Yulia and Bettencourt, Jesse and Duvenaud, David K},
  journal={Advances in neural information processing systems},
  volume={31},
  year={2018}
}

@inproceedings{zhao2023probabilistic,
  title={Probabilistic safeguard for reinforcement learning using safety index guided gaussian process models},
  author={Zhao, Weiye and He, Tairan and Liu, Changliu},
  booktitle={Learning for Dynamics and Control Conference},
  pages={783--796},
  year={2023},
  organization={PMLR}
}

@article{wang2024pre,
  title={Pre-trained Gaussian processes for Bayesian optimization},
  author={Wang, Zi and Dahl, George E and Swersky, Kevin and Lee, Chansoo and Nado, Zachary and Gilmer, Justin and Snoek, Jasper and Ghahramani, Zoubin},
  journal={Journal of Machine Learning Research},
  volume={25},
  number={212},
  pages={1--83},
  year={2024}
}

@article{marrel2024probabilistic,
  title={Probabilistic surrogate modeling by Gaussian process: A review on recent insights in estimation and validation},
  author={Marrel, Amandine and Iooss, Bertrand},
  journal={Reliability Engineering \& System Safety},
  pages={110094},
  year={2024},
  publisher={Elsevier}
}

@article{jin2024machine,
  title={Machine learning coffee price predictions},
  author={Jin, Bingzi and Xu, Xiaojie},
  journal={Journal of Uncertain Systems},
  volume={17},
  number={04},
  pages={2450023},
  year={2024},
  publisher={World Scientific}
}

\end{document}